\documentclass[journal]{IEEEtran}
\usepackage{amsmath,amssymb,amsfonts}
\usepackage{algorithmic}
\usepackage{graphicx}
\usepackage{textcomp}
\usepackage{color}
\usepackage{subfigure}

\usepackage[ruled,linesnumbered]{algorithm2e}
\usepackage{array}

\newtheorem{definition}{Definition} 
\begin{document}
\title{Optimal Multilayered Motion Planning for Multiple Differential Drive Mobile Robots with Hierarchical Prioritization
(OM-MP)}

\author{Zong Chen, Songyuan Fa, Yiqun Li*,
\thanks{This work was supported in part by the National Natural Science Foundation of China No. 51905185, National Postdoctoral Program for Innovative Talents No. BX20180109.

Authors are with the State Key Laboratory of Intelligent Manufacturing Equipment and Technology, School of Mechanical Science and Engineering, Huazhong University of Science and Technology, Wuhan, 430074, China. (e-mail: {\tt\small liyiqun@hust.edu.cn}).}
}

\markboth{Journal of \LaTeX\ Class Files,~Vol.~14, No.~8, August~2015}%
{Shell \MakeLowercase{\textit{et al.}}: Bare Demo of IEEEtran.cls for IEEE Journals}

\maketitle

\begin{abstract}
  We present a novel framework for addressing the challenges of multi-Agent planning and formation control within intricate and dynamic environments. This framework transforms the Multi-Agent Path Finding (MAPF) problem into a Multi-Agent Trajectory Planning (MATP) problem. Unlike traditional MAPF solutions, our multilayer optimization scheme consists of a global planner optimization solver, which is dedicated to determining concise global paths for each individual robot, and a local planner with an embedded optimization solver aimed at ensuring the feasibility of local robot trajectories. By implementing a hierarchical prioritization strategy, we enhance robots' efficiency and approximate the global optimal solution. Specifically, within the global planner, we employ the Augmented Graph Search (AGS) algorithm, which significantly improves the speed of solutions. Meanwhile, within the local planner optimization solver, we utilize Control Barrier functions (CBFs) and introduced an oblique cylindrical obstacle bounding box based on the time axis for obstacle avoidance and construct a single-robot locally aware-communication circle to ensure the simplicity, speed, and accuracy of locally optimized solutions. Additionally, we integrate the weight and priority of path traces to prevent deadlocks in limiting scenarios. Compared to the other state-of-the-art methods, including CBS, ECBS and other derivative algorithms, our proposed method demonstrates superior performance in terms of capacity, flexible scalability and overall task optimality in theory, as validated through simulations and experiments.
\end{abstract}


\IEEEpeerreviewmaketitle

\section{Introduction}
\label{sec:introduction}
\IEEEPARstart{A}{utumous} multi robot systems have gained widespread adoption in modern warehousing and manufacturing industries, particularly within sectors like E-commerce, ports and logistics. These systems play a pivotal role in facilitating efficient goods transportation, swift distribution of production components and automated material handling. For instance, in warehouse settings, autonomous robots are deployed to swiftly locate and transport goods to their designated stations, leading to noteworthy time savings and enhanced productivity. Similarly, within manufacturing environments, these robots execute tasks such as assembly line transport, material handling, and product inspection with remarkable precision and speed. The seamless functioning of these autonomous multi-robot systems hinges upon their ability to navigate the environment via the optimal paths, while skillfully circumventing potential collisions with both stationary and moving obstacles. To realize this objective, sophisticated path planning and obstacle avoidance algorithms are harnessed, which take into account the intricate dynamics of the system and the surrounding environment, ensuring efficient and safe navigation.
\par

\begin{figure}[!t]
    \centerline{\includegraphics[width=\columnwidth]{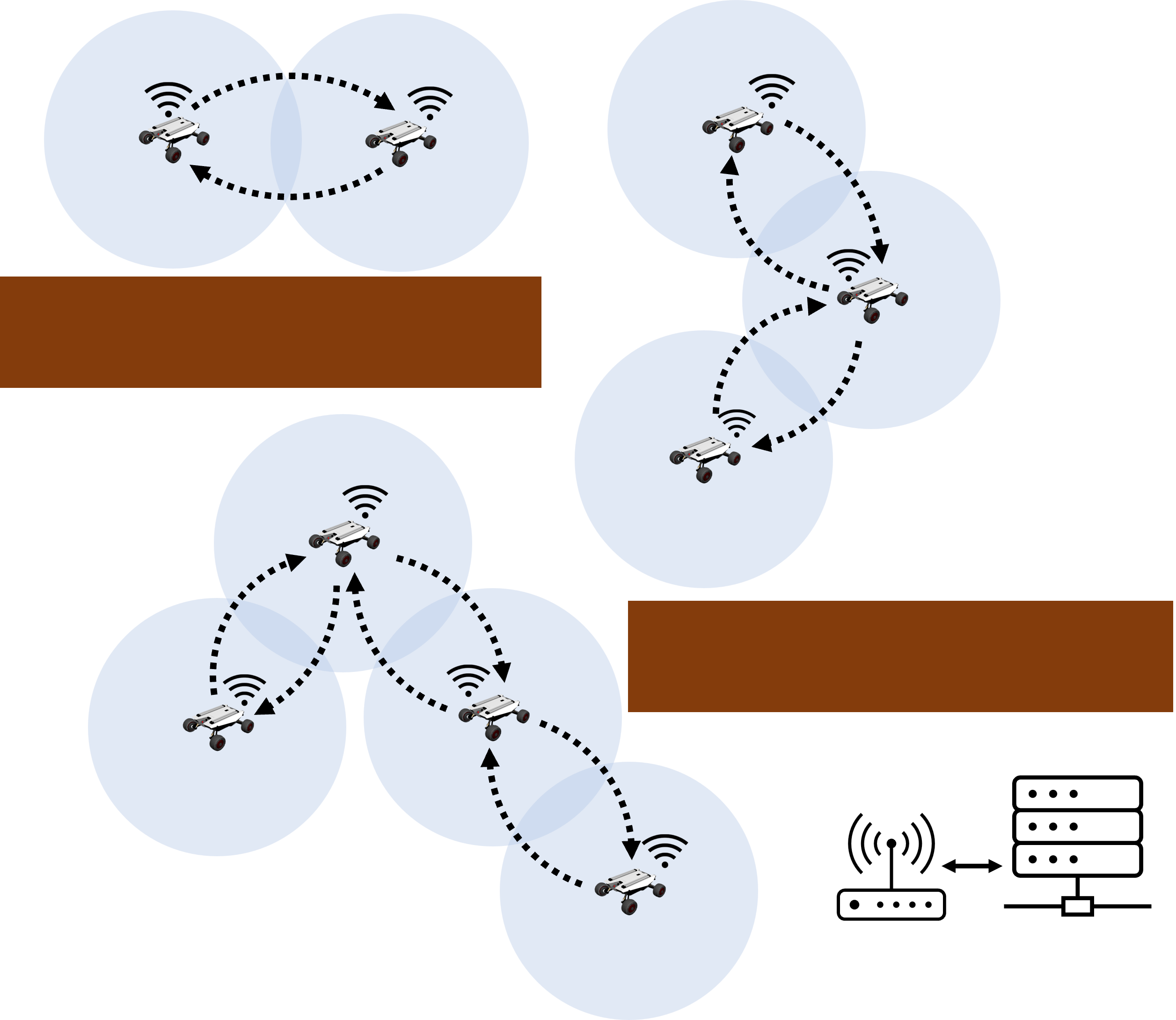}}
    \caption{Interactions between different robots in radius and communication with host and others.}
    \label{fig1}
\end{figure}
\par
Our proposed method optimizes the entire process including robot trajectory and control signal, phase by phase during the robot moving. Given a map of the environment and a set of robots with parameters such as the initial position, orientation, speed limitation, and goal information, during global path planning, AGS based on A*, Dijkstra, etc. is utilized to obtain a global feasible solution for a single robot, and incorporate additional strategies to ensure that different robot move to approach the goal, respectively. For local motion planning, we take into account the robot's kinematic model, dynamic obstacle avoidance, and environmental constraints to generate control signal to the robot's final actuators. Utilizing online trajectory generation to reduce the error and send the trajectory to the local and global planner in real-time, where the optimization speed of local planning can reach 20-30 Hz. To ensure the tracking effectiveness, we plan the trajectory for a future time domain $\Delta T$, but send only the most recent few time steps $\Delta t$ to the actuators, and the solving frequency of the local optimization is designed at 20 Hz for updating the control input. If solving failed or within the solution time, the subsequent $\Delta t$ of the previous solution sequence is used as a new control signal to be sent to the actuator. During each update period, the trajectory information obtained from the previous solution is sent to the lower-level actuators at a frequency of 200 Hz.
\par
Addressing nonlinear problems within an expansive and intricate robot space is a formidable challenge., these intricate issues often find representation through methods such as Mixed-Integer Programming \cite{mellinger2012mixed} or Sequential Convex Programming \cite{chen2015decoupled}. However, tackling this complicated problem across the entire feasible space for every single robot is both arduous and time-consuming, in this work, we adopt a different strategy, opting to decompose the overarching robot planning process into numerous single-robot trajectory planning problems. This approach involves harmonizing robot priorities, effectively sidestepping the potential complexity explosion associated with an escalating number of robots, but it also leads to difficulties in converging the final global optimal solution. The local planner that concentrates only on the environmental context surrounding $R$ each individual robot (as shown in Fig~\ref{fig1}, the azure circle). More specifically, we address the local trajectory planning by solving Non-Linear Programming (NLP) problems, thus deriving optimal solutions for each robot within its local environment. For more complex environment, we introduce prioritization and replanning strategy to ensure converge to global optimal solution.

\subsection{Related works}
\par
To address the Multi-robot path or trajectory planning problem, Sharon et al. introduced the collision-based constraint tree in \cite{sharon2015conflict}, which utilizes conflicts as the A* bi-level search cost function to discover optimal paths within a raster map. However, a limitation arises from the blind nature of the lower-level search, leading to exponential time complexity escalation as the number of robots increases. In efforts to mitigate time consumption while maintaining solution quality, alternative techniques have emerged \cite{barer2014suboptimal,li2021eecbs,chen2021symmetry}, these methods involve modifying the cost function or assigning higher weights to nodes with reduced likelihood of collision, and employing graph search-based group optimization strategies to alleviate problem complexity. Furthermore, the utilization of the right-of-way rule is employed to further trim computational demands in \cite{li2020efficient}. However, these approaches predominantly solving the planning tasks of all single robots as a whole, and execute computations on a central unit, failing to fundamentally address the challenge of centralized programming overhead and its associated computational complexity. Centralized programming focuses on comprehensive constraint consideration and get the global optimal solution, albeit at the expense of time efficiency, scalability of multiple robots and the flexibility to assign tasks on the fly. A more promising avenue lies in decentralized distribution methods that hold the potential to significantly curtail computational intricacies and time expenditure. In \cite{luis2019trajectory,luis2020online}, a decentralized distributed model predictive control (DMPC) scheme for multi-robot scenarios is employed, effectively dispersing global multi-robot planning into numerous subproblems. Within this framework, each robot adheres to the shortest path until it encounters an obstacle, at which point avoidance becomes a consideration. While this technique accelerates problem resolution, it can introduce deadlocks within intricate environments, especially when confronted with a substantial number of robots. Dedicated obstacle avoidance for obstacles by constructing a control barrier function (CBF) has been validated in a wide range of work \cite{desai2022clf,jankovic2018control,srinivasan2020control}. During the multi-robot system, \cite{wang2017safety} et al. constructed a $QP$ problem that guarantees system stability and obstacle collision free by using the Control Lyapunov Function (CLF) and the control barrier function, which has a better performance in dealing with obstacles.
\par
Multi-robots planning can be conceptualized as a challenge of steering multiple robots away from collisions across various scenarios and arrive its goal position, respectively. Likewise, every individual collision avoidance instance can be perceived as a conflict resolution puzzle within the context of Connected Automated Vehicles (CAVs) navigating through intersections, changing lanes and meeting.
\par
Navigating through intersections rapidly, orderly and securely with complex traffic flows is an extremely challenging task, especially within intricate scenarios marked by a plethora of obstacles. To tackle this challenge, the concept of  conflict zones has been integrated into prior methodologies, as evident in \cite{dresner2008multiagent, dresner2005multiagent, zhang2018decentralized}. These approaches have effectively addressed conflict avoidance quandaries in instances where dynamic obstacles like other robots, or vehicles are relatively sparse. However, in scenarios characterized by densely packed obstacles, these methods tend to mirror the First in, First out (FIFO) principle, which hinders the enhancement of the potential transportation efficiency. To overcome these limitations and elevate solution efficiency and quality, a collaborative group-based strategy has emerged, as exemplified in \cite{xu2019grouping}. In comparison to approaches like \cite{luis2020online,luis2019trajectory}, this method involves grouping robots (robots, AGVs or vehicles) with the same local objectives into platoons, treated as singular entities in planning. Nevertheless, it's important to note that this method's efficacy may be somewhat limited when robots with identical local goals are scarce, as its impact on efficiency improvement may not be as pronounced.
\par 
In addition to the aforementioned relatively mature and applied methods, there have been attempts to solve the multi-robots routing problem using the swarm intelligence \cite{di1998adaptive,wu20113d,teodorovic2003transport} and game theory with reinforcement learning \cite{nowe2012game}. However, these methods are often fallen into a local optimal. Besides, In complex environments with human-machine mixing M.C. et al. \cite{mavrogiannis2019multi} based on topology and inference, observing human behavior and inferring human behavioral intentions multi-robots for planning and navigation between complex and dense crowded environments. Deep Reinforcement Learning can achieve excellent planning and navigation performance in the training environment \cite{fan2020distributed}, but the migration learning for the scene requires additional training, and the uncertainty in the training process, it is difficult to ensure that in a variety of complex environments can quickly converge to the optimal feasible solutions.
\par
Taking the aforementioned factors into account, this paper introduces an optimized multilayered framework designed for multi-robot trajectory planning and control. The primary objective of this framework is to tackle the challenges that current methods encounter, balancing trade-off in algorithm time complexity, solution quality, framework capacity, scalability and the flexibility to assign tasks on the fly. The global planner of the framework focuses on single robot priorities and offers an initial solution, albeit that it might not be optimal or feasible for the whole process. On the other hand, the local planner adheres to the principles of priority and group consultation, resolve the nonlinear programming issue within a 3D space-time.
\subsection{Our Contributions}

The main contribution of this work is the proposal of an optimal multilayered framework with hierarchical priority and local group consultation for multi-robots trajectory planning, navigation, and control. This method can be applied to mobile robots, Automated Guided Vehicles (AGVs) in warehousing systems, and traffic scheduling at intersections. The specific contributions of this work are as follows:
\subsubsection{Multilayered Optimal Planning Framework}
\par
The proposed framework comprises two main levels, effectively decoupling the multi-robots trajectory optimization problem and improving solution efficiency. The global planner focuses on the collective path planning for all robots by supplying initial values to each robot using a combination of AGS and Non-Uniform Rational B-Splines (NURBS) \cite{piegl1996nurbs}. Additionally, it manages the priority relationships among the robots, thereby optimizing the overall time efficiency of the task. Meanwhile, the local planner runs in each individual robot, which primary tasks including dynamic obstacle avoidance, group consultation and transmitting control instructions to the robot's actuators.

\subsubsection{Hierarchical prioritization in the local planner}
\par
Hierarchical prioritization is assigned to each robot  based on its designated destination. This approach serves multiple purposes: it reduces the intricacy of subproblems, and enables the lower-priority robot to explore second-best solutions, these collective efforts drive the overall solution towards to the global optimum. Furthermore, the utilization of a hierarchical priority strategy effectively mitigates the occurrence of deadlocks in local planning processes.

\subsubsection{Cylinder obstacle avoidance with rank in 3D space-time}
\par
Within the local planner solution space, dynamic obstacles exhibit time-varying characteristics. Nevertheless, the solution of the local nonlinear programming solver is solely the discrete coordinates of the robot in order to balance the solving efficiency. To address this disparity, a novel approach involves the utilization of oblique cylinders for encapsulating obstacles like Fig~\ref{fig2} (c), which can wrap the continuous state between any two state points inside a slanting cylinder, as opposed to employing discrete spheres or positive cylinders as shown in Fig~\ref{fig2} (a) and Fig~\ref{fig2} (b). This innovative method is employed to construct a convex body obstacle space conducive to effective avoidance strategies.
\par
While incorporating the temporal dimension $t$ into 3D planning provides a certain degree of flexibility, the imposition of stringent constraints on obstacles across all planned trajectory points contributes to heightened solution intricacy. In light of this, the integration of a hierarchical obstacle avoidance approach offers a potential solution for mitigating this complexity. The weighting factors associated with obstacles and trajectory points are visually represented in Fig~\ref{fig2} (c), (d) and formula \ref{distribution_weight}.
\par
\subsubsection{Safe corridors by AGS}
\par
When robot navigates within complex environments like warehouses, obstacle avoidance becomes imperative. In contrast to the utilization of barrier expansion radii, the AGS method generates secure corridors. These corridors ensure pathfinding through areas that are relatively safe and devoid of static obstacles, enhancing the overall navigation safety.
\par
To address this concern, we proposed AGS method, which focuses on creating approximate secure corridors that envelop the fitted curve, 
the details are shown in $\uppercase\expandafter{\romannumeral3}.C$. This strategic modification ensures a safer distance from obstacles and optimizes path planning in such scenarios.

\subsubsection{Global planner with replanning}
\par
The initial values provided by the global planner to the local planner may not be globally optimal or feasible duo to the dynamic obstacles and the other robots. However, it can still yield a short path for an individual robot by AGS. To achieve a closer approximation to the global optimum and circumventing collision-prone zones or congested areas that demand prolonged traversal times, an additional replanning component has been incorporated into the global planner. This dynamic replanning mechanism aims to iteratively approach an improved solution in response to evolving conditions.

\section{Preliminaries}
\par
In this section, we delve into the robot model employed in our simulations and experiments, and also elaborate on the obstacle avoidance and NURBS curve methods that constitute in our approach. Furthermore, Table \ref{table} is included to furnish an overview of the parameters utilized for both simulation and experimental robots.

\begin{table}
    \caption{PARAMETERS OF SIMULATION AND EXPERIMENT ROBOT}
    \label{table}
    \setlength{\tabcolsep}{3pt}
    \begin{tabular}{|p{0.17\columnwidth}<{\centering}|p{0.35\columnwidth}<{\centering}|p{0.17\columnwidth}<{\centering}|p{0.17\columnwidth}<{\centering}|}
        \hline
        Symbol       &
        Description  &
        Value        &
        Unit           \\
        \hline
        $l $         &
        wheelbase    &
        0.52         &
        $\mathrm{(m) }$           \\
        $r$          &
        wheel radium &
        0.0875       &
        $\mathrm{(m)}$            \\
        $l_w$        &
        axle track   &
        0.45         &
        $\mathrm{(m)}$            \\
        $m$          &
        weight       &
        25.0         &
        $\mathrm{(kg)}$           \\
        $v_{max}$    &
        max speed    &
        2.0          &
        $\mathrm{(m/s)}$          \\
        \hline

    \end{tabular}
    \label{tab1}
\end{table}

\begin{figure}[h]
    \centerline{\includegraphics[width=0.8\columnwidth]{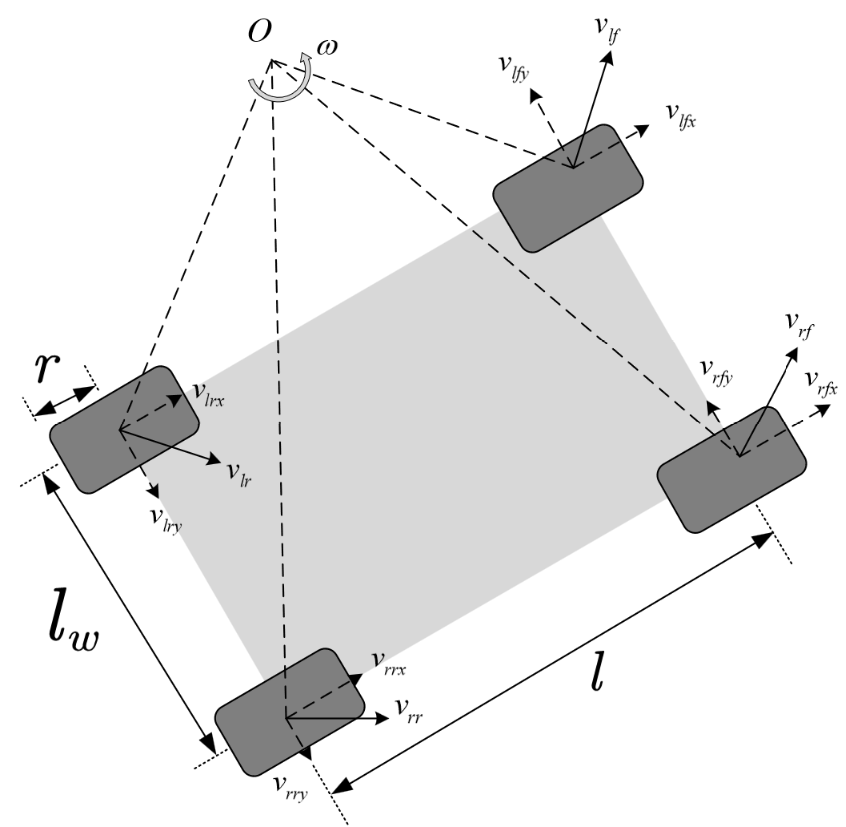}}
    \caption{Illustration of pose and control variables of the robot.}
    \label{fig3}
\end{figure}

\subsection{Robot Model}
The configurations of the robot are presented in Fig~\ref{fig3}, Denote the configuration space of the robot as $Q=SE\left( 2 \right) \times S^1\times S^1$ with local coordinates $\mathrm{q_{0}} =\,\,\left[ x,y,v_r,v_l,\theta \right] $, where $\left[ x,y,\theta \right] $ are the position and orientation of the robot, $\left[ v_r,v_l \right] $ are the speed of right and left wheels,  $\left[ u_r,u_l \right] $ are the angular acceleration of right wheel and left wheels. In this way, the robot model can be expressed as follows:
\begin{equation}
    \label{agent_model}
    \begin{aligned}
        &\dot{x}\,\,  =\,\,\frac{v_l+v_r}{2}\,\,\cos \theta \\
        &\dot{y}\,\,=\,\,\frac{v_l+v_r}{2}\sin \theta \\
        &\dot{v}_r\,\,=\,\,r\,\,{ u_r}                 \\
        &\dot{v}_l\,\,=\,\,r\,\,{u_l}\\
        &\dot{\theta}\,\,=\,\,\frac{v_r-v_l}{l_w}
    \end{aligned}
\end{equation}
\par
where ${r}$ is the wheel radius, ${l_w}$ is the axle track.
\par
In configuration space, we map the 2D space (ignored the robot height) to the 3D space with time, we choose the $q=\left[ x,y,v_r,v_l,\theta,t \right]^T $ as the state variables, and choose the $U=\left[ u_r,u_l \right]^T $ as control variables, so the state-space equation can be expressed as:
\par
\begin{equation}
    \label{state_trans}
    \dot{q}={Aq+BU}
\end{equation}
where $A\in \mathbb{R}^{6\times 6}\,\,$ is the state-transition matrix, and $B\in \mathbb{R}^{6\times 2}$ is the Control input matrix.
\par
Moreover, the system~\ref{state_trans} can be rewritten to the affine form as,
\begin{equation}
    \label{affine_dynamics}
    \dot{s}=f\left( s \right) +g\left( s \right) u
\end{equation}
where the state $s\in q \subset \mathbb{R}^6$ and control $u \in U \subset \mathbb{R}^2$, $f$ and $g$ are locally Lipschitz continuous.\par
The details of system~\ref{state_trans} and \ref{affine_dynamics} are given in Appendix \ref{app:A}. 

\subsection{Control Barrier Functions}
\begin{definition}
    (extended class $\mathcal{K} _{\infty}$). A continues function $\alpha :\left( -b,a \right) \rightarrow \mathbb{R} $ for $a,b >0$ is called extended class $\mathcal{K} _{\infty}$ function if:\par
    $\bullet$ it is strictly increasing;\par
    $\bullet$ it is s.t. $\alpha(0) = 0$;\par
    $\bullet$ $\underset{r\rightarrow -\infty}{\lim}\alpha \left( r \right) =-\infty ,\underset{r\rightarrow \infty}{\lim}\alpha \left( r \right) =\infty $;
\end{definition}

\begin{definition}
    \label{safe_set_def}
    (Forward invariance) \cite{chai2018forward}. For a set $\mathcal{C} \subset \mathbb{R} $ is called forward invariant for system~\ref{affine_dynamics}, if $s_0\in \mathcal{C} \,\,$ and $s\left( t,s_0 \right) \in \mathcal{C} ,\forall t\geqslant 0$.
\end{definition}
Consider a set $\mathcal{C}$, which defined as the $zero-superlevel\,\,set$ of a continuously differentiable function $\mathcal{B}:\mathcal{D}\subset \mathbb{R} ^n\rightarrow \mathbb{R}$, yielding:\par
\begin{equation}
    \begin{aligned}
        \label{safe_set}
        \mathcal{C} &= \left\{ s  \in \mathcal{D}\subset \mathbb{R}^n : \mathcal{B}(s) \geq 0 \right\} \\
        \partial (\mathcal{C}) &= \left\{ s\in \mathcal{D}\subset \mathbb{R}^n : \mathcal{B}(s) = 0 \right\} \\
        \text{Int}(\mathcal{C}) &= \left\{ s\in \mathcal{D}\subset \mathbb{R}^n : \mathcal{B}(s) > 0 \right\}
        \end{aligned}        
\end{equation}
and, we called the $\mathcal{C}$ as the safe set.\par
The time derivative of $\mathcal{B}(s)$ along the state trajectories is \par
\begin{equation*}
    \frac{d\mathcal{B} \left( s \right)}{dt}=\nabla \mathcal{B} \left( s \right) \dot{s}=\frac{\partial \mathcal{B} \left( s \right)}{\partial s}\dot{s}=\frac{\partial \mathcal{B} \left( s \right)}{\partial s}\left( f\left( s \right) +g\left( s \right) u \right)
\end{equation*}
and using the Lie derivative as the formation
\begin{equation*}
    \frac{d\mathcal{B} \left( s \right)}{dt}=\mathcal{L} _f\mathcal{B}\left( s \right) +\mathcal{L} _g\mathcal{B}\left( s \right) u
\end{equation*}

\begin{definition}
    (Control Barrier Functions (CBFs)). For a continuous and differential function $\mathcal{B}(s)$, and a set $\mathcal{D}$, if there exists an extended class $\mathcal{K} _{\infty}$ function $\alpha$ such that
    \begin{equation*}
       \underset{u\in U}{sup}\left\{ \mathcal{L} _{f}\mathcal{B}\left( s \right) +\mathcal{L} _g\mathcal{B}\left( s \right) u+\alpha \left( \mathcal{B}\left( s \right) \right) \geqslant 0 \right\} 
    \end{equation*}   
    for all $s \in \mathcal{C}$, where $\mathcal{C} \subset \mathcal{D} \subset \mathbb{R}^n$, and $\mathcal{C}$ satisfied the equation~\ref{safe_set} for system \ref{affine_dynamics}.
\end{definition}
We refer to function $\mathcal{B}(\cdot)$ as the control barrier function of system~\ref{affine_dynamics}.

\subsection{Obstacle avoidance}
\begin{figure}[t]
    \centering
    \subfigure[Positive cylindrical bounding box]{
        \includegraphics[scale=0.25]{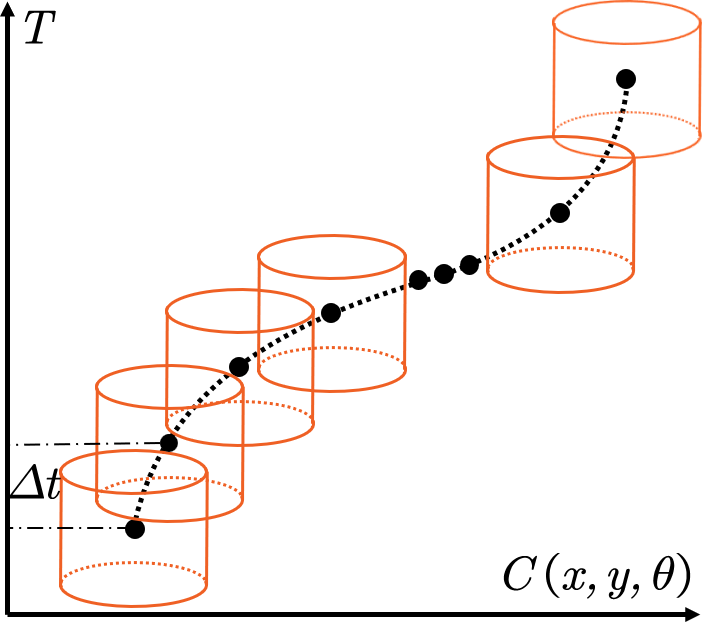}
        \label{obs_a}
    }
    \quad
    \subfigure[Sphere bounding box]{
        \includegraphics[scale=0.25]{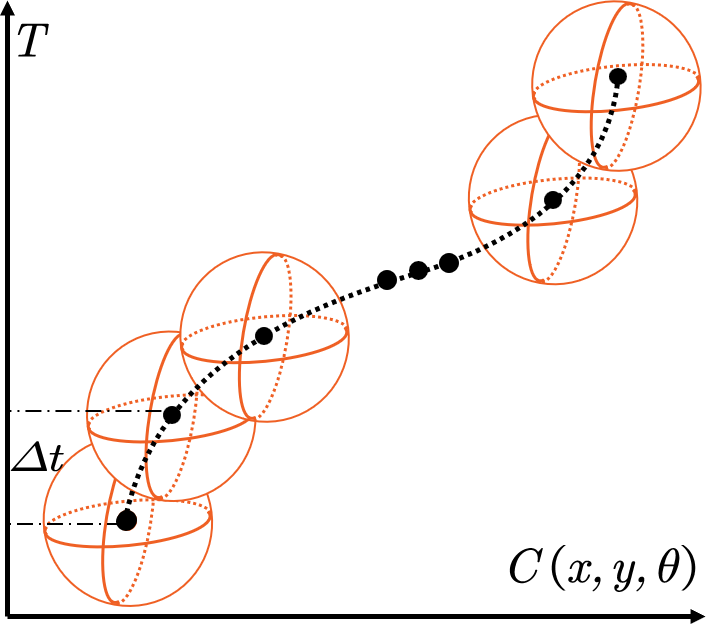}
        \label{obs_b}
    }
    \quad
    \subfigure[Oblique cylindrical bounding box]{
        \includegraphics[scale=0.195]{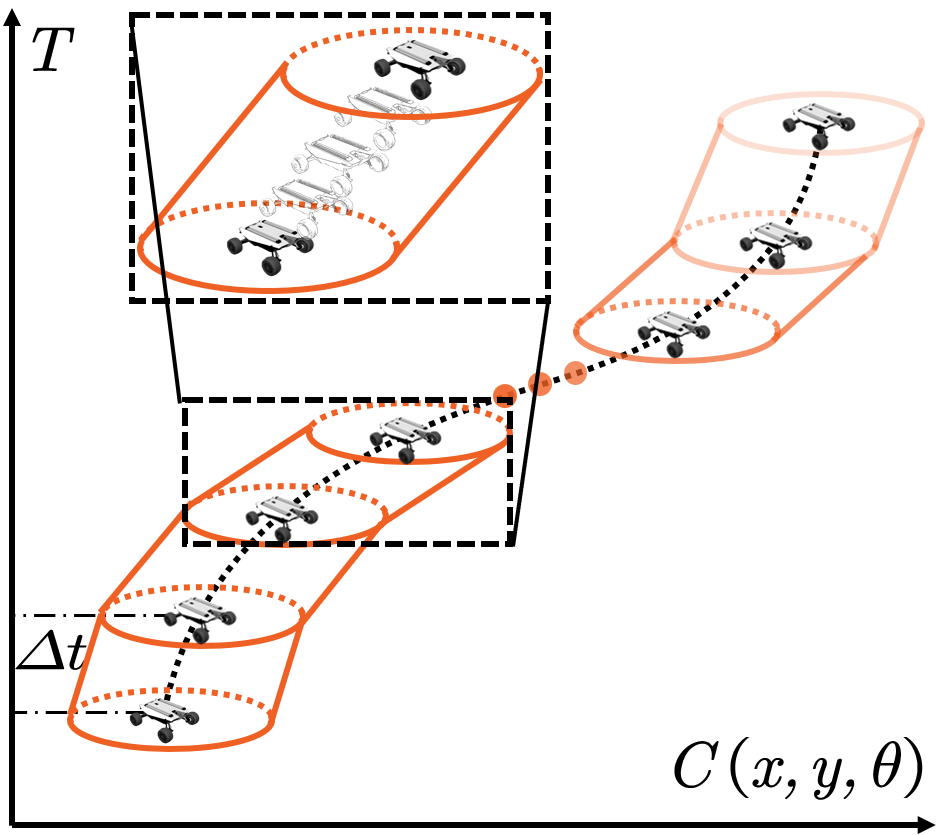}
        \label{obs_c}
    }
    \quad
    \subfigure[Distribution of weighting factors for local planning]{
        \includegraphics[scale=0.215]{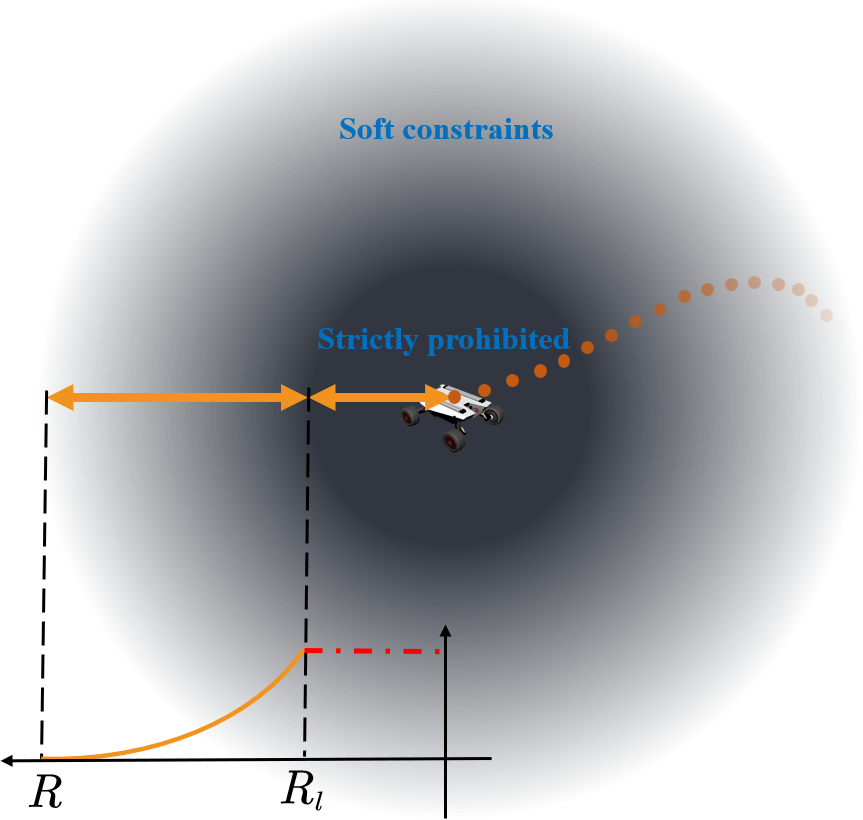}
        \label{obs_d}
    }
    \quad
    \caption{Schematic illustration of a trajectory covered by bubbles bounding box that represent a local free space corridor in (a), (b) and (c), (d) is the mapping distribution of obstacle and planning trajectory point weights coefficients in 2D. }
    \label{fig2}
\end{figure}

In the local planner solution space, dynamic obstacle avoidance are continuous and time-varying. However, the solution provided by the local planner is discrete, and the obstacle constraints incorporated are also in discrete form. In previous studies, attempts were made to depict obstacle variations over time using methods such as spheres or positive cylinders \cite{lopez2017predictive}, as shown in Fig~\ref{obs_a} and Fig~\ref{obs_b}. While, this approach often leads to a simplistic representation of obstacles and consumes additional solution time and spatial resources. The positive cylinder method assumes that obstacles or the robot itself remain static at discrete point positions for a certain time interval before and after $\frac{\varDelta t}{2}$. Yet, this assumption doesn't align with the reality of potentially time-varying obstacles and robot itself. Instead, we adopt an alternative approach by approximating the position and state variations between two discrete matching points linearly, by connecting the discrete coordinates of the same obstacle at different spatio-temporal, constructing continuous convex spaces for each pair of adjacent obstacle spatio-temporal coordinates. The obstacle avoidance challenge is thus transformed into the robot position to line segments distance problem, within a 3D spatial-temporal. By enveloping the robot's conceptual space by oblique cylindrical bounding box, we establish connections between these continuous convex body, ultimately yielding a navigable path free of obstacles.
\par
To attain a local optimal solution and generate trajectories that are not only feasible and optimal but also incorporate avoidance measures within a 3D spatial-temporal. This process takes into account robot-specific constraints like speed, acceleration, and mechanical structure. Importantly, addressing the avoidance aspect can be succinctly articulated as follows:

\begin{equation}
    \boldsymbol{d}\in R^{n_N\times m_{obs}}\,\,>\,\,d_0
    \label{distan}
\end{equation}
where $\boldsymbol{d}$ is the sequence of distances between each solution match point and all the obstacles in the 3D spatial-temporal, and ${d_0}$ is the safety distance between the robot and obstacle. 
\par
In this work, the obstacle avoidance is achieved by the control barrier function $\mathcal{B}(s)$. For every single predicted trajectory point of planning, the safe set $\mathcal{C}$ can be combined by the $\boldsymbol{d}$, therefore $\mathcal{B}(s)$ is designed as $\mathcal{B}(s) = \boldsymbol{d}_{ij} - d_0$.
\par
The extended class $\mathcal{K} _{\infty}$ function shows how fast the states of the system approach to the boundary of safe set $\mathcal{C}$, In practice, the extended class $\mathcal{K}_{\infty}$: $\alpha(\cdot)$, is often designed by the linear combination of a positive real number $\gamma$ and Barrier function $\mathcal{B}(\cdot)$. In this work, the extended class $\mathcal{K}_{\infty}$ is designed as :$\alpha(\mathcal{B}(s)) = \gamma\mathcal{B}(s) $ with $\gamma >0$.     (The details to choose $\alpha(\cdot)$ is given in Appendix \ref{app:B}).\par
By design the CBF, the obstacle avoidance is achieved in the optimization as hard constraints such that
\begin{equation}
    \label{CBFS}
    \mathcal{L} _{f}\mathcal{B}\left( s \right) +\mathcal{L} _g\mathcal{B}\left( s \right) u+ \gamma\mathcal{B}\left( s \right)  \geqslant 0 
\end{equation} 
\par
Moreover, the current position of the robot as the initial value of the local trajectory. Therefore, obstacles near the distal portion of the trajectory do not need to be treated as hard constraints, as this would unnecessarily increase the computational complexity of the underlying planning process. As depicted in Fig~\ref{obs_c} and Fig~\ref{obs_d}, obstacles are added as hard constraints of the planning within close to the robot's current position $R_l$ by CBF, and obstacles are constrained by penalizing the distance between the obstacles and the points along the planning trajectory using a weighted factor that decreases in proximity within $R_l$ to $R$ as described by Equation \ref{distribution_weight}.

\begin{equation}
    \varOmega =\left\{ \begin{array}{c}	\infty ,\,\,    \left\| r \right\| <R_l\\	\frac{c_0}{c_1r+c_2},\,\,  R_l\leqslant \left\| r \right\| \leqslant R\\\end{array} \right. 
    \label{distribution_weight}
\end{equation}
where $c_0$, $c_1$ and $c_2$ represents the coefficients parameters.
\par
\par

\begin{figure*}[t]
    \centerline{\includegraphics[width=1.5\columnwidth]{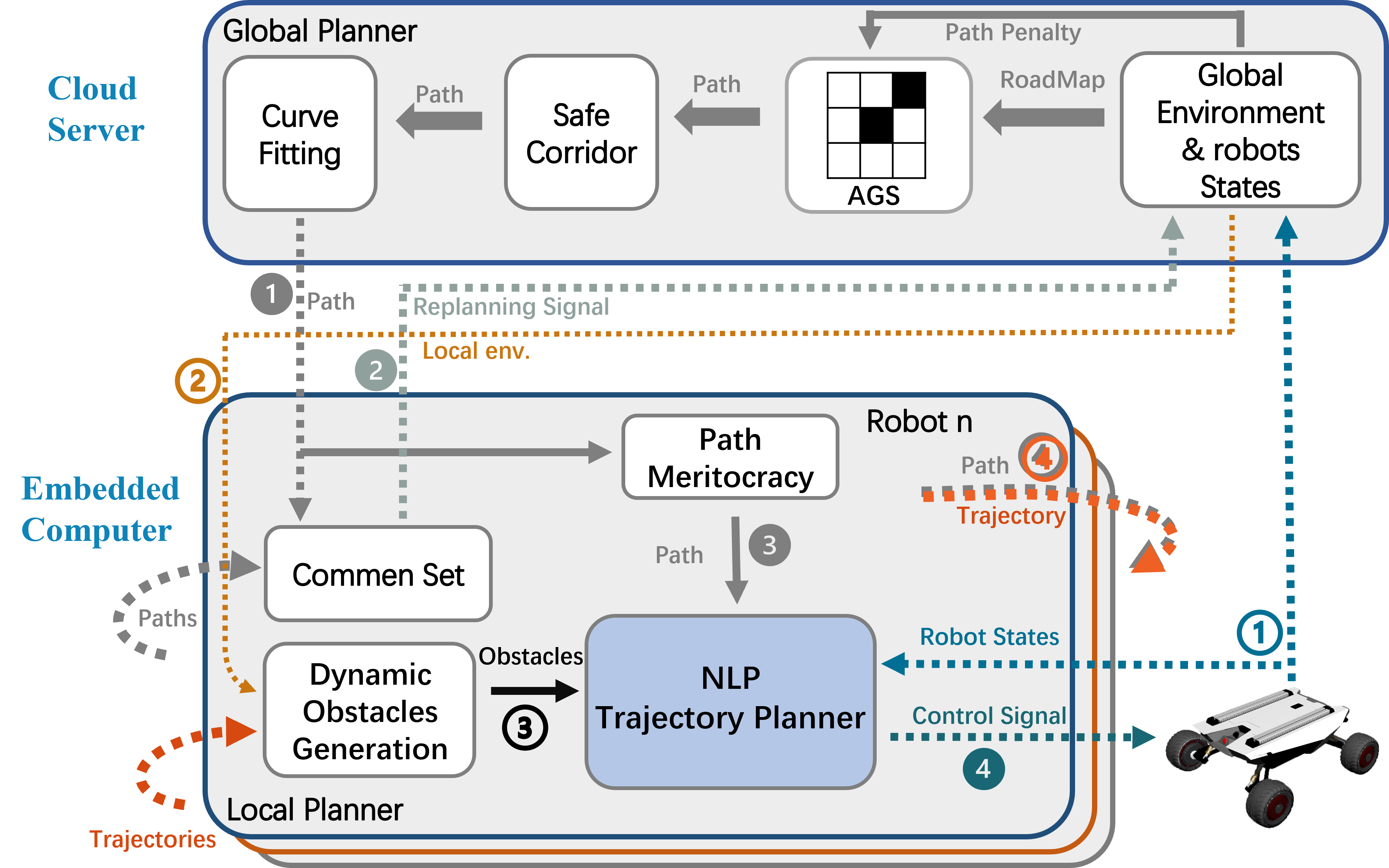}}
    \caption{The overview of this framework. Where the hollow serial numbers depict the sequential passage of signals from the global planner to the execution terminal, while solid serial numbers signify the sequential feedback of signals from the terminal back to the planner. And the solid and dotted lines indicate the wired and wireless transmission of the signal, respectively.}
    \label{fig5}
\end{figure*}

\section{Methodology}
In this part, we will present the main algorithms in this work: the global planner with replan, the local trajectory planner with negotiate, Augmented Graph Search (AGS) methods and the replanning algorithm.
\subsection{Assumption}
The trajectory generated should obey the following criteria:
\begin{itemize}
  \item [1)] 
  All robots with collision-free initial and goal configuration.
  \item [2)]
  The static obstacles $\mathcal{O} $, the start and end positions of robot are given as prior knowledge. The dynamic obstacles and other robots are initially unknown, but their positions, velocities and local trajectories can be obtained for the future time horizon by the global planner when the robot is within the negotiate circle.
  \item [3)]
  All robots share their positions, global paths and local trajectories to the global planner in real time.
  \item [4)]
  The global planner calculates the distance of each robot, and send the other robots number within the negotiate circle in real-time.
\end{itemize}

\subsection{Multilayer Optimal framework}
The comprehensive architecture of the multilayer optimal framework is vividly illustrated in Fig~\ref{fig5}. In this diagram, the "Common Set" component detects whether the current path and paths of higher priority robots share a common set, which in turn triggers the replanning signal. The "Path Meritocracy" module, operates within the local planner horizon, determining the selection between new global path and the original path. The comprehensive algorithm for the Multilayer Optimal framework is provided as Algorithm~\ref{algorithm_Overview}.
\par
Our method consists of the following steps.
\par
\begin{algorithm}
    \caption{OM-MP Algorithm}\label{algorithm_Overview}
    \KwData{grid-based map $\mathcal{F} ^{\mathcal{G}}$, initial and goal position
    ${P_{*,s}}$\ ${P_{*,g}}$, replan trigger ${r_{trig}}$, AGS function $\mathcal{G} _{\mathcal{F}}\left(  \cdot\right) $,
    replan AGS function $\mathcal{R} _{\mathcal{G} ,\mathcal{F}}\left( \cdot \right) $ }
    \KwResult{Control signal $\mathcal{C} _{sign}$}
    \par
    \uIf {not First plan} {
        $\boldsymbol{S} \gets \mathcal{G} _{\mathcal{F}}\left( \mathcal{F} ^{\mathcal{G}}, {P_{*,s}}, {P_{*,g}}\right) $
    } \ElseIf {not ${r_{trig}}$} {
        $S _i\gets  \mathcal{R} _{\mathcal{G} ,\mathcal{F}}\left( \mathcal{F} ^{\mathcal{G}}, {P_{n,s}}, {P_{n,g}} \right) $
        \par
        update $\boldsymbol{S} $
    }
    \par
    $\mathcal{S} \gets R_{env}\left( pose,R \right) $
    \par
    $\mathcal{O}_{dy} \gets \mathcal{S}$  \par
    $\mathcal{O} \gets (\mathrm{O}_{stat},\mathcal{O}_{dy})$ 
    \par
    \tcp*[h]{Create a new thread to solve NLP}
    $T_{new}\,\,=\,\,Thread_{nlp}\left( \mathcal{O} ,S_{start},S_{goal},P_{ref} \right) $
    \par
    \uIf {$Thread_{nlp}$} {
        $T_l\gets T_{new}$
    } 
    \Else{$T_l\gets T_{old}(N_p,end)$}
    $T_{old} \gets T_l$
    \par
    $\mathcal{C} _{sign} \gets T_l$
    \par
    publish $\mathcal{C} _{sign}$ to Robots 
    \par
\end{algorithm}
1. The global planner gets the initial starting points $P_{*,s}\in \mathbb{R}^{N\times 3}$ and end points $P_{*,g}\in \mathbb{R}^{N\times 3}$ for all robots, within the planning space $\mathcal{F} ^{\mathcal{G}}$. Employing AGS algorithms, it initially acquires global solution $\boldsymbol{S}_0\subset \mathbb{R}^{N\times (V+1)}$, where ${N}$ represents the robot count and ${V+1}$ indicates the trajectory points and weights (However, solution $\boldsymbol{S}_0$ may unfeasible during the whole process for the robots, due to the dynamic environment is not considered). Following this, the global planner conveys $S_0\in \boldsymbol{S}$ to the respective robots.
\par
2. The robot receives the initial path ${S_0}$ and employs it as the temporary goal and initial solution for the local planner to perform motion planning. Subsequently, it forwards the control signal $\mathcal{C} _{sign}\subset \mathbb{R}^{U_C\times N_P}$ to the final actuators, here, ${U_C}$ signifies the number of control values, and ${N_P}$ indicates the time steps of future control values. Simultaneously, the robot's sensors transmit its states ${p}$ to the local and global planner, and the local planner re-evaluates motion planning within the robot's designated radius ${R}$. Notably, the local planner serves a dual role as both a planner and a tracker throughout the process.
\par
Furthermore, the local planner publishes a portion of the global path of the future time domain, $P_{g,t}$ (where, $g$ means the global path, $t$ domains the $t$ time in the future) that it has received from global planner, as well as the local trajectory $T_{l,t}$ (where $l$ means the local trajectory) that has been obtained by solving the nonlinear programming problem to other robots within the radius ${R}$. Meanwhile, it transmits the robot's position and state values, along with the replan signal, to the global planner.
\par
3. Upon receipting the robot states, the global planner computes the sequence of neighboring robots $\mathcal{S}$ situated within the designated radius ${R}$. These robots are treated as dynamic obstacles and are subsequently communicated to the respective robot. Furthermore, if the replan signal is activated, the global planner initiates the replanning procedure, strategically avoiding the regions previously occupied and planned by robot, and the newly derived path is then transmitted to the corresponding robot.
\par
4. The local planner observes the local environment through the robot's perception of its neighboring robots denoted by $\mathcal{S}$, and evaluates whether other robots within its vicinity share a common set of actions. If a common set is identified, the local planner modifies the trigger of the replanning signal and prompts a request for a new path from the global planner.
\par
5. Upon receiving a new global path, the local planner compares the costs between the newly provided path and the original path leading to the terminal, and choose the fast one to follow. It's important to note that the pass time of the original trajectory generated with older path could be elongated within the Nonlinear Programming (NLP) solver duo to the lower priority, resulting in waiting. 
\par
6. At the terminal of the robot, the local planner subscribes to both global and local information within ${R}$, denoted as $P_{g,t}$ and $T_{l,t}$ respectively, originating from other robots within the $\mathcal{S}$ set, which conducted according to their respective weights, acting as the planning priorities. Then, the local planner views the trajectory points of the other robots with higher priority as dynamic obstacles within the solution space, effectively mitigating the complexity of the solution process. It's worth noting that the weights represent the generalized distance from the current position of the robot to the goal position,  and they are dynamically updated in real-time.
\par
\begin{figure}[h]
    \centerline{\includegraphics[width=\columnwidth]{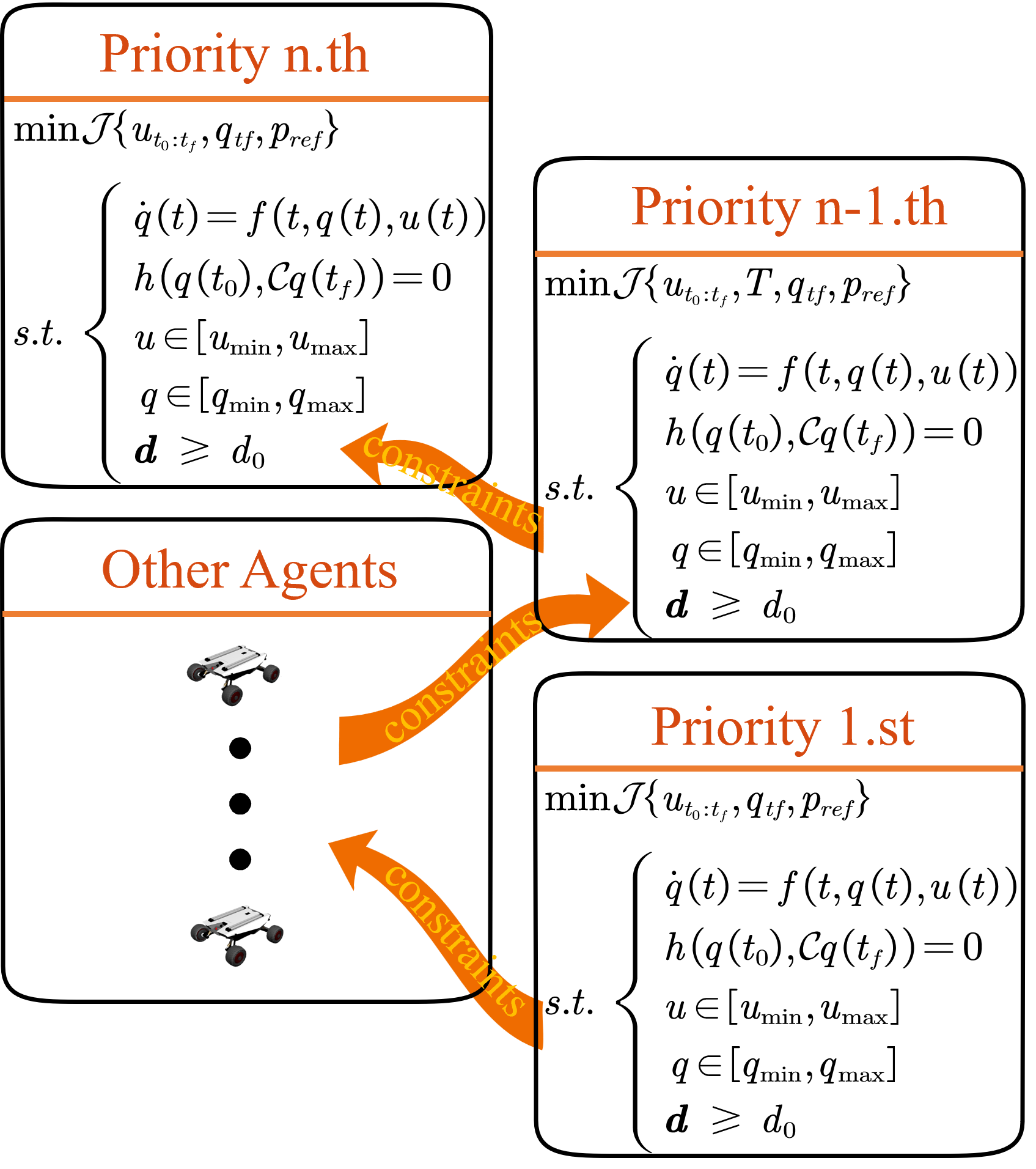}}
    \caption{Illustration of the priority order, the orange arrows stand for the trajectory points of higher priority robots, which are considered as dynamic obstacles by lower priority ones.}
    \label{Priority_order}
\end{figure}

Within the radius $R$, the local planner undertakes trajectory planning, during this process, robots with higher priority are planned first, and their trajectory points are conveyed to robots of subsequent priority level as dynamic obstacles. This sequence continues iteratively for all other robots of lower priority, and the details of this approach are visually depicted in Fig~\ref{Priority_order}.

\par
The algorithm overview is shown in Algorithm~\ref{algorithm_Overview}.
\par
\subsection{Augmented Graph Search (AGS) method}
\par
\begin{figure}[t]
    \centering
    \subfigure[Simple $1$-order graph search]{
        \includegraphics[width=0.45\columnwidth]{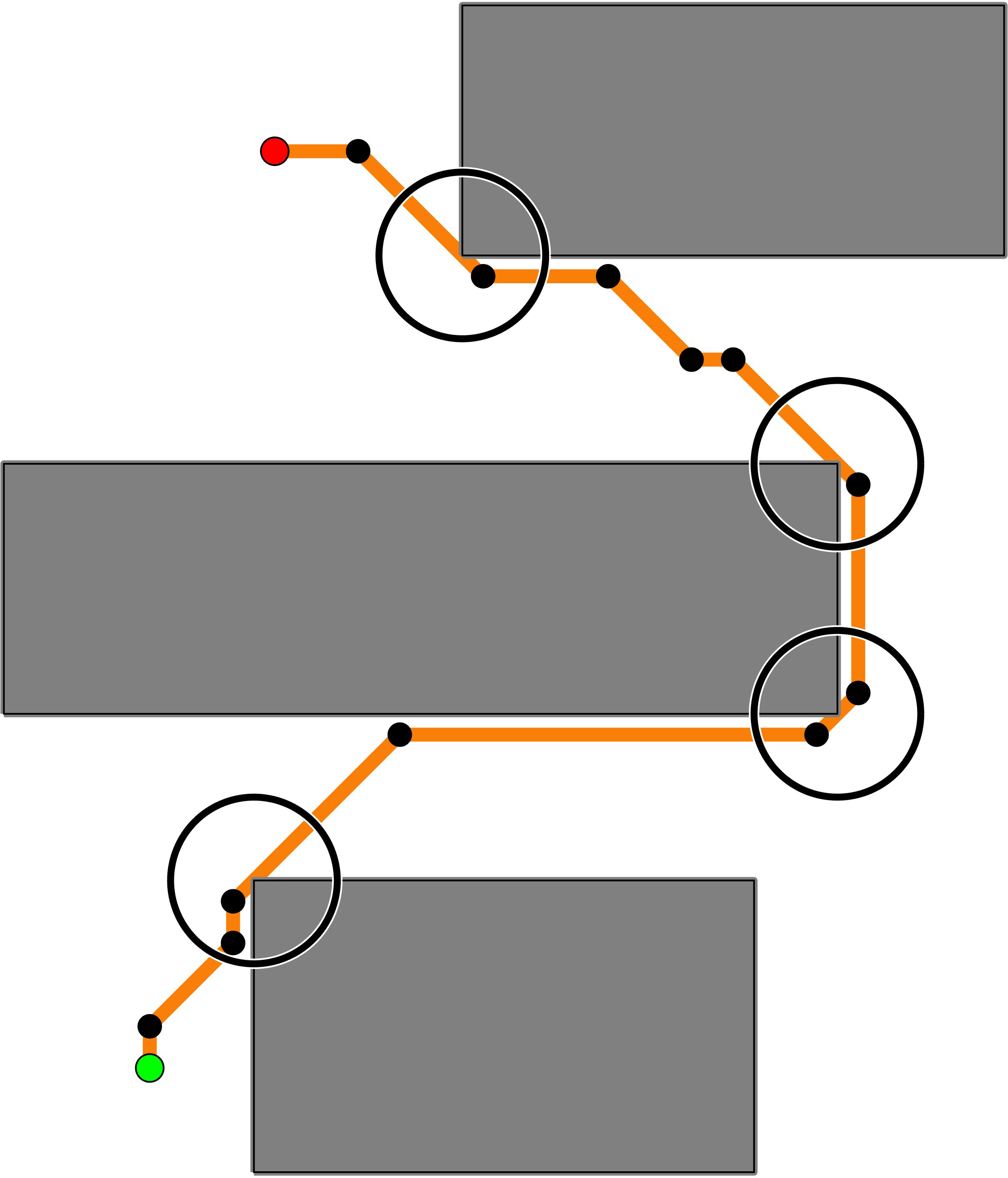}
        \label{a_star}
    }
    \quad
    \subfigure[Augmented Graph Search]{
        \includegraphics[width=0.43\columnwidth]{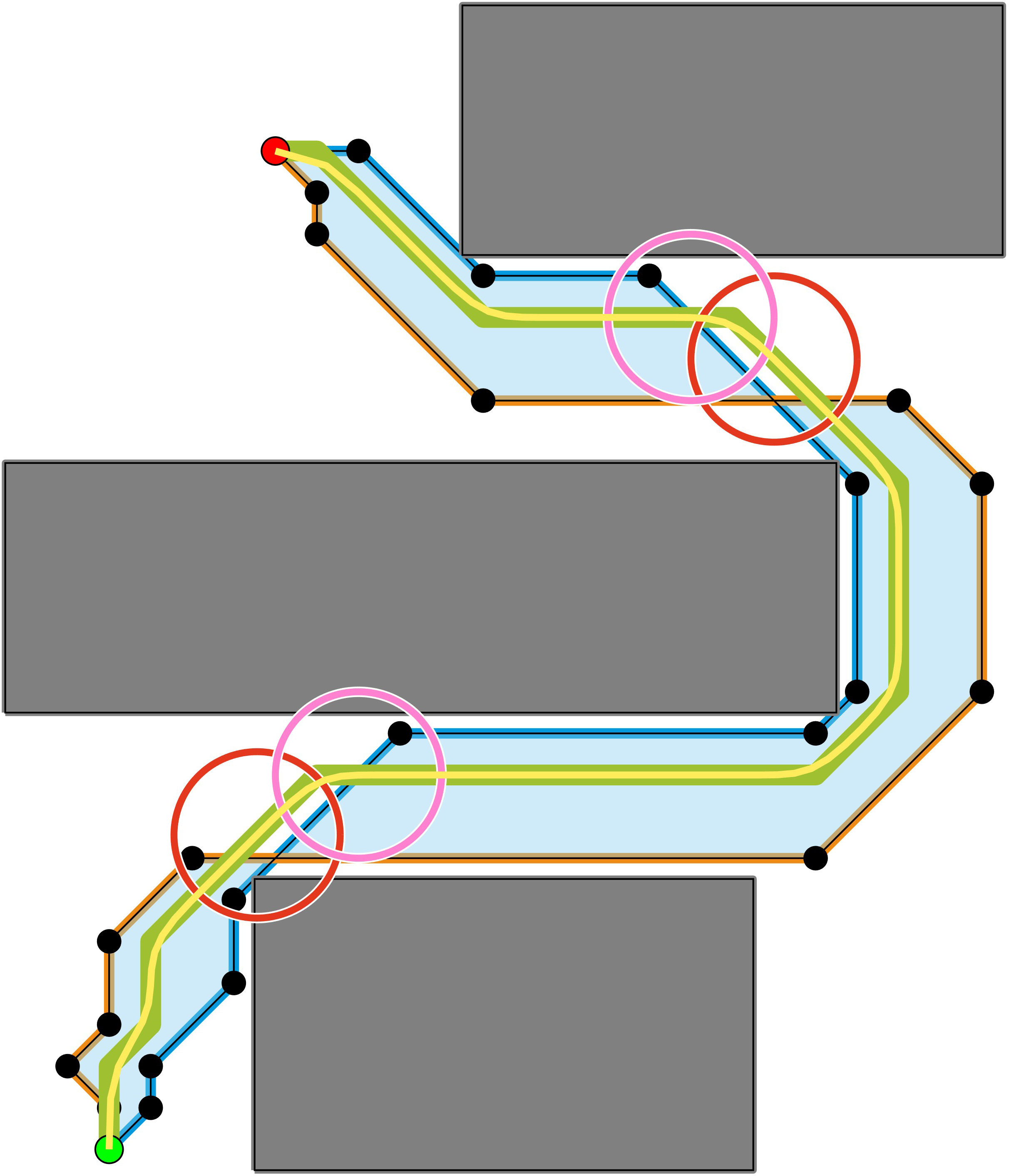}
        \label{enhanced}
    }
    \caption{Comparison between $1$-order AGS and $N$-Order AGS is illustrated. The gray blocks denote obstacles, while the green and red points signify the starting and target points within the graph search. In (a), an orange line depicts a simple search path. In (b), the blue and brown lines correspond to paths resulting from a $1$-order and $N$-order search, respectively. The green line represents a path generated during the $1-N$ order graph search process. Additionally, the yellow line represents a fitted curve derived from the green path. Moreover, black dots indicate penalty points. Within this graph, the $4$-order enhanced graph path is represented by the brown line, while the $2$-order enhanced graph path is depicted by the green line. }
    \label{Enhanced graph search}
\end{figure}

The use of a simple graph search method ($1$-order AGS) sometimes leads to the paths that come too close to obstacles, particularly at points of inflection, as indicated by the black circle in Fig~\ref{a_star}. On the contrary, the AGS method is capable of forming approximate secure corridors that around the fitted curve, as depicted in Fig~\ref{enhanced}. This envelopment is such that the area enclosed by the paths generated by the $1$-order and $N$-order AGS algorithms nearly entirely covers the fitted curve. It is crucial to note that while the curve fitted within the red-circled area in Fig~\ref{enhanced} extends beyond the safe corridor established by the first to $N$-order AGS algorithms, it still can be proved that this region remains collision-free, obviously.
\par
The blue line depicted in Fig~\ref{enhanced} represents the path generated by a 1st-order AGS, which can be obtained through basic graph search techniques like A* or Dijkstra. However, simple graph searches exhibit limitations, particularly in terms of generating corners near the top points of obstacles. To address this issue, we identify all the turning points (shown as black points) along the curve of previous orders, as demonstrated in Fig~\ref{Enhanced graph search}, these turning points are treated as additional obstacles in the higher order of AGS, and the new curve avoids these identified areas. Notably, the curve from the lower order is positioned closer to obstacles in the direction of the target point, by penalizing the inflection points along the curve of the lower order, the higher order's curve is inclined to veer away from obstacles. The interplay between higher and lower order curves culminates in the creation of a safety space, denoted by the light blue area in Fig~\ref{enhanced}, As a result, although the fitted curve within the red circle does not lie entirely within the safety space, its safety is assured.
\par
Even with second-order or higher-order curves, the tendency to adhere to a lower-order (1st) curve persists.
As higher-order curves are incorporated, the security of retaining lower-order inflection points as obstacles that far more away from the obstacles becomes evident. Consequently, the introduction of higher-order curves leads to the creation of new inflection points in advance, aimed at avoiding these virtual obstacle points. This phenomenon is illustrated by the pink circles in Fig~\ref{enhanced}. As a result, higher-order curves are positioned further away from obstacles compared to their lower-order counterparts.
\par
In Fig~\ref{enhanced}, the $1$st order AGS curves (blue), $N$th order curves (brown), and the intermediate curves spanning from $1$st to $N$th order (green) are all obtained by AGS, thus ensuring their paths remain free of obstacles. Nonetheless, when fitting collision-free segments with curves, the potential emergence of new collision points becomes apparent \cite{li2023apollo,chen2024fdspc}. For all points requiring fitting with curves, an iterative process is initiated. Continuous Collision Detection (CCD) \cite{pan2012collision} is employed to sequentially assess the collision-free nature of the path curve between two fitted points [$t_i$, $t_{i+1}$]. In the event a collision point is detected, an additional fitting point $t_{i'}$ is introduced at the midpoint between the two original AGS points along the path. If collisions persist, additional fitting points are incrementally incorporated. Notably, due to the inherent characteristics of fitting curves, the newly integrated fitting points only affect the adjacent two fitting segments. Consequently, these additions do not introduce new variations or collisions within the pre-existing sections.
\par
However, the strategy of generating safety space through AGS does have limitations, particularly in situations involving narrow passages or when grid points are substantially larger than the environment itself. Moreover, challenges could arise when obstacles are situated within areas surrounded by multiple feasible paths. Despite these constraints, this approach remains valuable for identifying safety zones on a segment-by-segment basis, ultimately producing safety curves.
\subsection{Global path planner with replan}
The primary role of the global path planner is to find all feasible paths, provide them to robots in real-time, and update them by broadcasting at predefined intervals. Additionally, the global path planner is responsible for re-finding paths in response to activation of a robot's replan signal $r_{trig}$. After receiving the start points ${P_{*,s}}$ (where $*$ represents all points and $s$ denotes the start point) and goal points ${P_{*,g}}$ (where $g$ signifies the goal point) of all robots, the planner utilizes AGS method to compute the global paths for each robot individually.
  To approximate optimal solutions, the replanning function is integrated into this part. When the global planner receives the replanning trigger ${r_{trig}}$, it evaluates both the robot's number and position that initiated the signal. Subsequently, it generates a new path by implementing the replanning algorithm $\mathcal{R} _{\mathcal{G} ,\mathcal{F}}$. This integration ensures the ability to avoid the deadlocks in narrow and crowded environment, guiding the robot towards the global optimal solution.
\par
\begin{algorithm}
    \caption{Global path planner}\label{algorithm1}
    \KwData{The number of robot ${N}$, grid-based map $\mathcal{F} ^{\mathcal{G}}$, initial and goal position
    ${P_{*,s}}$, ${P_{*,g}}$, replanning trigger ${r_{trig}}$, AGS function $\mathcal{G} _{\mathcal{F}}\left(  \cdot\right) $,replan AGS function $\mathcal{R} _{\mathcal{G} ,\mathcal{F}}\left( \cdot \right) $, curve fitting function $\mathcal{S} _{fit}\left(  \cdot\right) $}
    \KwResult{ Sequence of path $\boldsymbol{S}$}
    \For{$i \leftarrow 0 $ \KwTo ${N}$}{
    $S_i \leftarrow$ $\mathcal{G} _{\mathcal{F}}\left(  \mathcal{F} ^{\mathcal{G}}, {P_{*,s}}, {P_{*,g}} \right) $\par
    $S_i\gets \mathcal{S} _{fit}\left( S_i \right) $ \par
    $\boldsymbol{S}\gets S_i$\
    }
    \While{global planner}{
    \If{trigger}{
    ${n \leftarrow }$ ${r_{trig}.num}$ 
    \par
    ${P_{n,s} \leftarrow r_{trig}.pos}$  
    \par
    $S_n \leftarrow$ $\mathcal{R} _{\mathcal{G} ,\mathcal{F}}\left( \mathcal{F} ^{\mathcal{G}}, {P_{n,s}}, {P_{n,g}}\right) $ \par
    $\boldsymbol{S}\gets S_n$
    }
    \For{$i \leftarrow 0$ \KwTo $N$}{
        \If{$\varXi _{tab}\left( i,1 \right) \leqslant R$}{
            $\mathcal{S} \gets i$
        }
        $\boldsymbol{S}\gets \mathcal{S} $
    }
    publish ${\boldsymbol{S}}$
    }
\end{algorithm}

After receiving feedback signals from the robots, the global planner computes the square Euclidean distance for each individual robot. These distances are then used to create a distance table $\varXi _{tab}$ in the global planner, which is updated whenever feedback changes. Assuming that the global planner has received a series of robots $\left[ \left( x_1,y_1 \right) ,\left( x_2,y_2 \right) ,\cdots \left( x_m,y_m \right) \right]$, represented as $X=\left[ x_1,x_2,\cdots x_m \right] ^{\mathrm{T}}$ and $Y=\left[ y_1,y_2,\cdots y_m \right] ^{\mathrm{T}}$, the distance table $\varXi _{tab}$ can be formulated as follows,
\par
\begin{equation}
    \begin{aligned}
        &\mathcal{X} =\left( \boldsymbol{X}+\boldsymbol{X}^{\mathrm{T}} \right) -2XX^{\mathrm{T}}\\
        &\mathcal{Y} =\left( \boldsymbol{Y}+\boldsymbol{Y}^{\mathrm{T}} \right) -2YY^{\mathrm{T}}
    \end{aligned}
\end{equation}
where $\boldsymbol{X}=X\odot X\cdot 1_{1\times \mathrm{m}}, \boldsymbol{Y}=Y\odot Y\cdot 1_{1\times \mathrm{m}}$, and $\odot$ is the Hadamard Product.
\par 
The table $\varXi _{tab}$ can be formulated as:
\par
\begin{equation}
        \varXi _{tab}=\mathcal{X}+\mathcal{Y}
\end{equation}
\par
The square Euclidean distance between each robot and other robots can be obtained by look up the robot number of a particular row or column of table $\varXi _{tab}$ in $O\left( 1 \right) $.
\par The global planner algorithm is presented in Algorithm \ref{algorithm1}.

\subsection{Local trajectory planning with negotiate}
The local motion planning algorithm is designed to solve the challenge of reaching the local terminal position while adhering to diverse constraints, encompassing safety and collision avoidance. The local planner acquires a series of robot numbers with higher priority from the global planner and considers them as dynamic obstacles. To streamline the trajectory generation process, only the distance within a radius of ${R}$ is taken into consideration. Additionally, planning priority rules are incorporated into the planning process to reduce complexity and prevent the deadlocks.
\par
\begin{figure}[h]
    \centering
    \subfigure[]{
        \includegraphics[width=0.42\columnwidth]{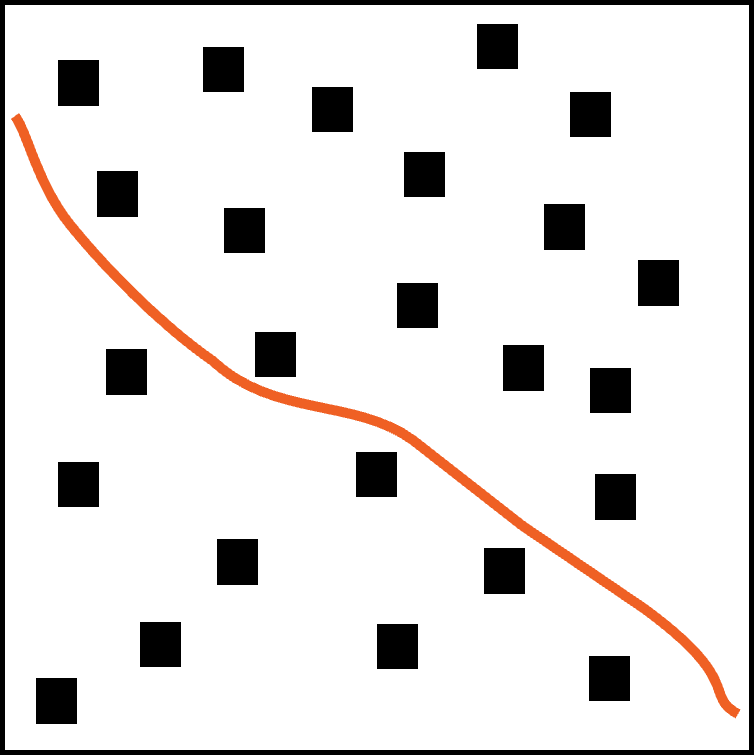}
    }
    \quad
    \subfigure[]{
        \includegraphics[width=0.42\columnwidth]{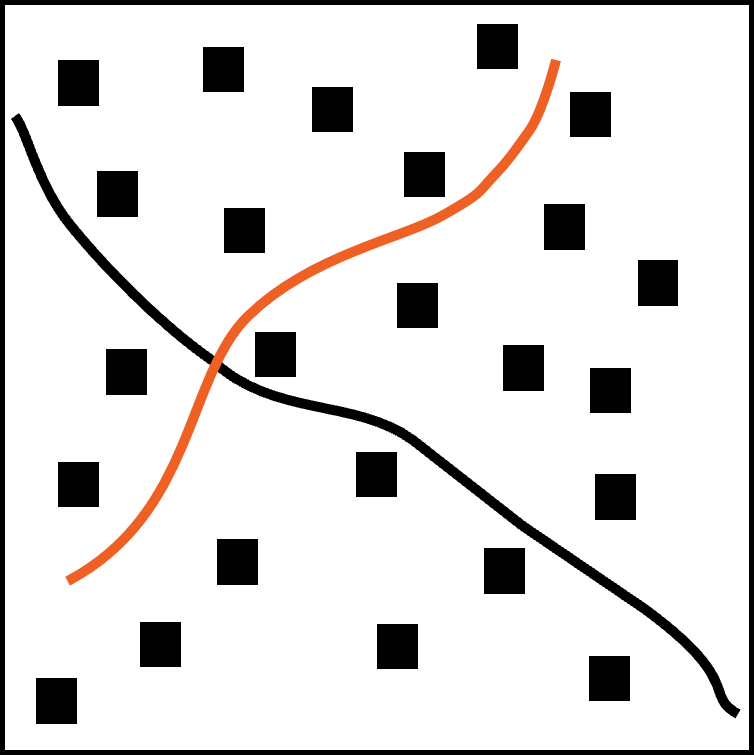}
    }
    \quad
    \subfigure[]{
        \includegraphics[width=0.42\columnwidth]{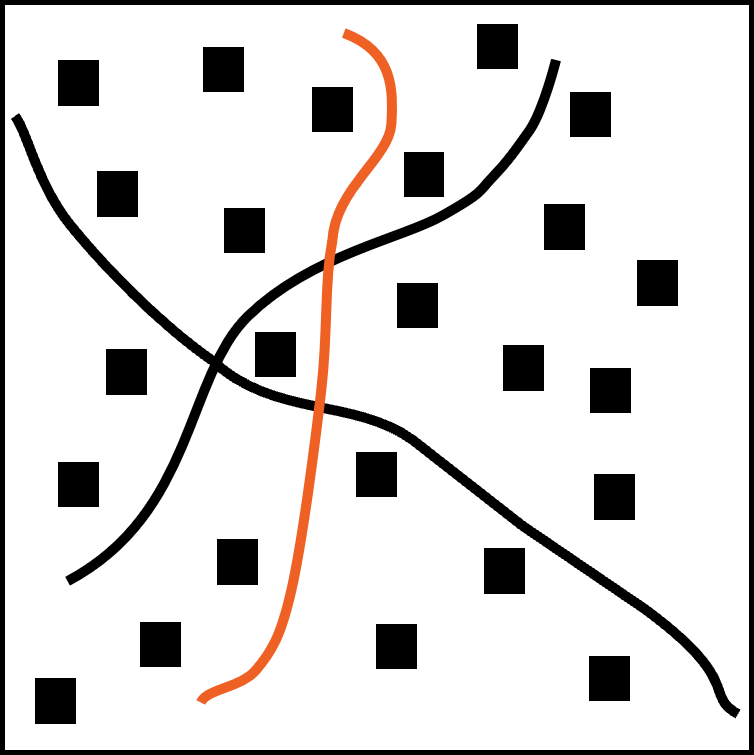}
    }
    \quad
    \subfigure[]{
        \includegraphics[width=0.42\columnwidth]{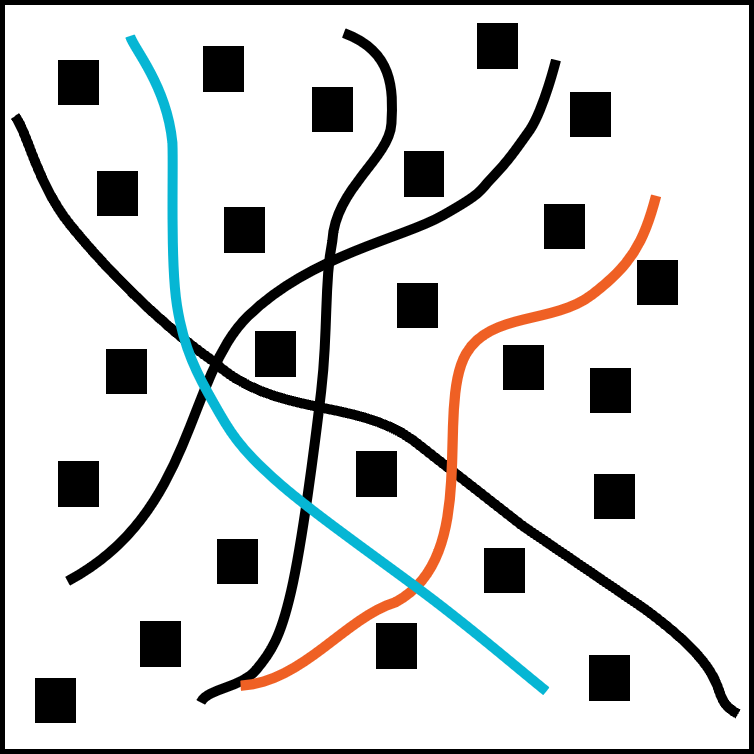}
    }
    \caption{The demonstration of the prioritized trajectory optimization in communication radius.}
    \label{fig6}
\end{figure}

As depicted in Fig~\ref{fig6}, robots with higher priority are planned first, followed by those with lower priority. Notably, lower priority robots take into account the trajectories of higher priority robots as dynamic obstacles.

\par
\par
Robots with higher priority than the current robot will be added to the local planner as dynamic obstacles, along with other constraints.  The local motion planner is outlined in Algorithm \ref{algorithm2}.

\begin{algorithm}[h]
    \caption{Local motion planner}\label{algorithm2}
    \KwData{Global path ${S}$, surrounding robots sequence ${\mathcal{S}}$, final actuator states ${p_{f}}$,
        surrounding robots global paths ${P_{g,*}}$ and their local trajectories ${T_{l,*}}$, local planning function ${L_{p}(\cdot)}$}
    \KwResult{Trajectory ${T_{l}}$}
    \For{${i \leftarrow 0 }$ \KwTo  ${\mathcal{S} .size }$}{
        \If{$weight \,\,<\,\,\mathcal{S} _i.weight $}{
            ${O \leftarrow \mathcal{S}_i}$ \par
            \For{${t \leftarrow 0 }$ \KwTo  ${P_{g,*} .size}$}{
                \If{$\left( P_{g,t}\,\,in\,\,\mathcal{S}_i \right) \cap \left( T_l\right) $}{
                    ${r_{trig}}\,\,=\mathrm{True}$ 
                }
            }
        }
    }
    \For{${i \leftarrow 0}$ \KwTo ${O.size}$}{
        ${\mathcal{O} \leftarrow T_{l,t}}$
    }
    ${P_s \leftarrow p_{f}}$ 

    ${T_l \leftarrow}$ ${L_{p} \left(\mathcal{O}, S, P_s \right)}$
    \par
\end{algorithm}
\subsection{Replanning algorithm}
The replanning algorithm is an improvement of AGS.
\par
The activation of the replan signal ${r_{trig}}$ prompts the global planner to initiate the replanning process. During this process, a new path is generated, considering not only the static obstacles within the map but also the regions that the robot will traverse in a future time domain $\varOmega$ along the old path. The replanning algorithmic for selecting between the original path and the newly computed global path is presented in Algorithm \ref{algorithm3}. To achieve this, the approach involves penalizing these regions $\varOmega$ by assigning elevated values within the heuristic function. Consequently, the AGS avoids exploring the same regions. Alternatively, a similar outcome can be achieved by incorporating the original regions into the CloseList. Both strategies work towards ensuring that the new path appropriately accounts for both the dynamic and static aspects of the environment.

\begin{algorithm}[h]
    \caption{Replanning algorithm}\label{algorithm3}
    \KwData{robot number ${n}$, grid-based map $\mathcal{F} ^{\mathcal{G}}$, robot present position ${P_{n,now}}$ and goal position ${P_{n,g}}$, curve fitting function $\mathcal{S} _{nurbs}\left( \cdot \right) $, robot surrounding radius ${R}$,
    AGS function $\mathcal{G} _{\mathcal{F}}\left( \cdot \right) $}
    \KwResult{ New path ${S}$}
    \par
    \For{${i \leftarrow \frac{R.size}{2} }$ \KwTo  ${R.size}$}{
        \For{${j \leftarrow \frac{R.size}{2}}$ \KwTo ${R.size }$}{
            ${\hat{h} \leftarrow 10000}$ 
        }
    }
    ${\mathcal{F} _{old}}^{\mathcal{G}}=\mathcal{F} ^{\mathcal{G}}$ \par
    $\mathcal{F} ^{\mathcal{G}} \leftarrow \hat{h}$ 
    \par
    $S_n \leftarrow$ $\mathcal{G} _{\mathcal{F}}\left( P_{n,now},P_{n,g},\mathcal{F} ^{\mathcal{G}} \right) $\par
    $S_{n}^{'}\gets \mathcal{S} _{fit}\left( S_n \right) $ \par
    $\mathcal{F} ^{\mathcal{G}}={\mathcal{F} _{old}}^{\mathcal{G}}$ 
    \par
    $\bar{v}=\frac{S_{passed}}{T_{costed}}$
    \par
    $T_{new}=\frac{S_{new}.size}{\bar{v}}$
    \par
    $T_{old}=T_{opti}+\frac{S_{remain}-S_{pred}}{\bar{v}}$
    \par
    \If{$T_{new}<T_{old}$}{
        $S\gets S_{n}^{'}$
    }
    \par
\end{algorithm}

\begin{table}
    \caption{Local Optimal Limits}
    \label{table2}
    \setlength{\tabcolsep}{3pt}
    \begin{tabular}{|p{30pt}<{\centering}|p{100pt}<{\centering}|p{30pt}<{\centering}|p{40pt}<{\centering}|}
        \hline
        Symbol                    &
        Description               &
        Value                     &
        Unit                        \\
        \hline
        $dis_f$                   &
        Forward search distance   &
        1.5                       &
        $\mathrm{(m)}$                         \\
        $v_{lim}$                 &
        Speed limit               &
        2                         &
        $\mathrm{(m/s)} $                      \\
        $t_{step}$                &
        Time step                 &
        3                       &
        $\mathrm{(m)}$                         \\
        $t_{lim}$                 &
        Time solution limit       &
        0.2                       &
        $\mathrm{(s)}$                        \\
        $N_{poi}$                 &
        Number of matching points &
        15                        &
        $/$                         \\
        $f_{sol}$                 &
        Solving frequency         &
        20                     &
        $\mathrm{(Hz)} $                       \\
        $f_{con}$                 &
        Control frequency         &
        200                       &
        $\mathrm{(Hz)}$                        \\
        $a_{limit}$                &
        Wheel acceleration limitation          &
        1.5                       &
        $\mathrm{(m/s^2)}$                         \\
        $s_{init}$                &
        Initial solution          &
        $/$                       &
        $/$                         \\
        \hline
    \end{tabular}
    \label{tab2}
\end{table}

\section{Local solver formulation}
\par
During each solving interval, the local planner tackles an optimization problem with various constraints. To improve the solver's efficiency, we refrain from fixing the final position within the local planning environment; instead, we treat it as a soft constraint within the problem. As a result, the primary components of the objective function are time, final position and path reference error. Building upon the aforementioned objective function and constraints, the local trajectory planning problem can be formulated as follows:
\par
\begin{subequations}
    \begin{align}
        \min \quad &\mathcal{J} \left\{t_f,u_{t_0:t_f},q_{t},p_{ref} \right\} \label{a}\\ 
        s.t.  \quad  &\dot{q}\left( t \right) =f\left( t,q\left( t \right) ,u\left( t \right) \right)  \label{b}\\   
        &h\left( q\left( t_0 \right) ,\mathcal{C} \cdot q\left( t_f \right) \right) =0  \label{c}\\ 
        &u\in \left[ u_{\min},u_{\max} \right]  \label{d}\\
        &q\in \left[ q_{\min},q_{\max} \right]  \label{e}\\   
        &\dot{\mathcal{B}}\left( s \right) +\gamma \mathcal{B} \left( s \right) \geqslant 0 \label{f}
    \end{align}
\end{subequations}
here \ref{b} denotes the system's kinematic constraints, while equality \ref{c} represent the initial and part of terminal restrictions of the system. The limits of the state and control values of the system are given by \ref{d} and \ref{e}, and the \ref{f} signifies the collision-free constraint.
\par
We use the Ipopt \cite{wachter2006implementation} to solve optimal problems, and differential is make by automated tools, such as CasADi \cite{Andersson2019}, CppAD \cite{bell2012cppad}, ADOL-C \cite{griewank1996algorithm}.




\section{Simulation \&\& Experiment}
This section introduces four scenarios that serve as demonstrations of our approach. The implementation was carried out using C++ within the ROS (Robot Operating System) environment and executed on a PC running Ubuntu 20.04. The PC itself is furnished with an AMD Ryzen 5600X CPU operating at 3.7GHz, alongside 16 GB RAM. In order to identify static obstacles, binary information is extracted from the map image utilizing OpenCV. The initial demonstrations were showcased using Rviz, and subsequently, the data underwent additional processing and visualization through MATLAB.
\par
In our simulation experiments, all robots have a communication interaction radius of ${R = 1.5m}$. The acceleration of the left and right wheels is limited to ${1.5 \,m/s^2}$, and the velocity is limited to  ${2,\ m/s}$.
\par
\subsection{Warehouse and Random scenarios}
\begin{figure*}[th]
    \centering
    \subfigure[OM-MP in warehouse scene planning]{
        \includegraphics[width=2\columnwidth]{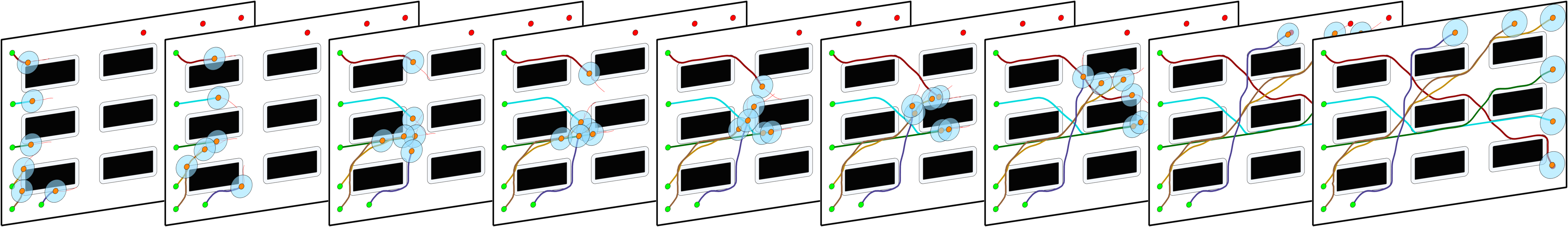}
        \label{ommp_a}
    }
    \quad
    \subfigure[OM-MP in random scene planning]{
        \includegraphics[width=2\columnwidth]{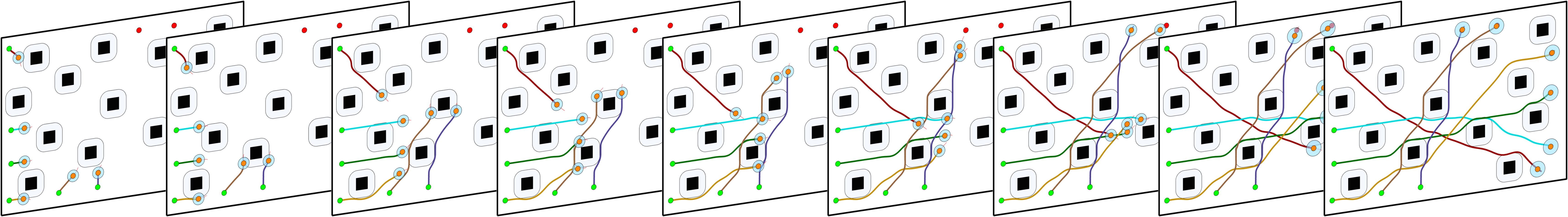}
        \label{ommp_b}
    }
    \caption{(a) and (b) correspond to the different planning processes of the OM-MP planning in a regular warehouse and a system with random obstacles respectively. Here, black color denotes the obstacles, while the green and red dots indicate the start and end points, the orange points represent the current position of the intelligent robots, and the red thin line indicates the current planning trajectory of the robot. Furthermore, the light blue circles plotted around the robot signify the range within which the robot establishes communication.}
\label{OM-MP Planning}
\end{figure*}

In the context of transporting goods within warehouse environments, coordinating and avoiding collisions among multiple robots is of paramount importance, the scenario is depicted in Fig~\ref{ommp_a}. To efficiently and optimally accomplish the task, a hierarchical prioritization strategy is introduced. This strategy involves assigning priority to robots that are further away from their destinations during the local avoidance process, which helps ensure an overall optimal outcome in terms of task completion.
\par
The planning system's resilience is tested in environments populated with random obstacles, as depicted in Fig~\ref{ommp_b}. A crucial aspect of our planning approach is the circumvention of static obstacles within the environment using AGS and fitting curves, ensuring that the generated paths adhere to the requirement for avoiding static obstacles. 
In cost function, the sub-item that closely corresponds to the fitted curve allocates a higher weight, as a result, nearby robots are inclined to decelerate or halt to avoid dynamic obstacles, rather than generating novel paths for evasion. This strategy not only ensures robust and stable when moving along the paths but also mitigates the issue of frequent changes in planned paths, which could complicate the process of accurate tracking.
\par

\begin{table}
    \caption{Computation time}
    \label{table3}
    \setlength{\tabcolsep}{4pt}
    \begin{tabular}{|p{40pt}<{\centering}|p{70pt}<{\centering}|p{50pt}<{\centering}|p{35pt}<{\centering}|}
        \hline
        Number of robots                    &
        AGS planning (s)               &
        Curve fitting (s)                     &
        Total solving (s)                        \\
        \hline
        1                   &
        0.0934   &
        0.0002                       &
        0.0936                         \\
        4                 &
        0.3358               &
        0.0199                         &
        0.3557                       \\
        8                &
        0.5763                 &
        0.0351                       &
        0.6114                         \\
        12                 &
        0.7550       &
        0.0500                       &
        0.8050                         \\
        16                 &
        0.9131 &
        0.0648                        &
        0.9778                         \\
        20                &
        1.0576          &
        0.0794                     &
        1.1370                        \\
        24                 &
        1.2515         &
        0.0958                       &
        1.3473                       \\
        28                &
        1.5066          &
        0.1148                       &
        1.6214                         \\
        32                &
        1.8154          &
        0.1371                       &
        1.9526                         \\
        \hline
    \end{tabular}
\end{table}
\par
\begin{figure}[h]
    \centering
    \subfigure[]{
        \includegraphics[width=0.445\columnwidth]{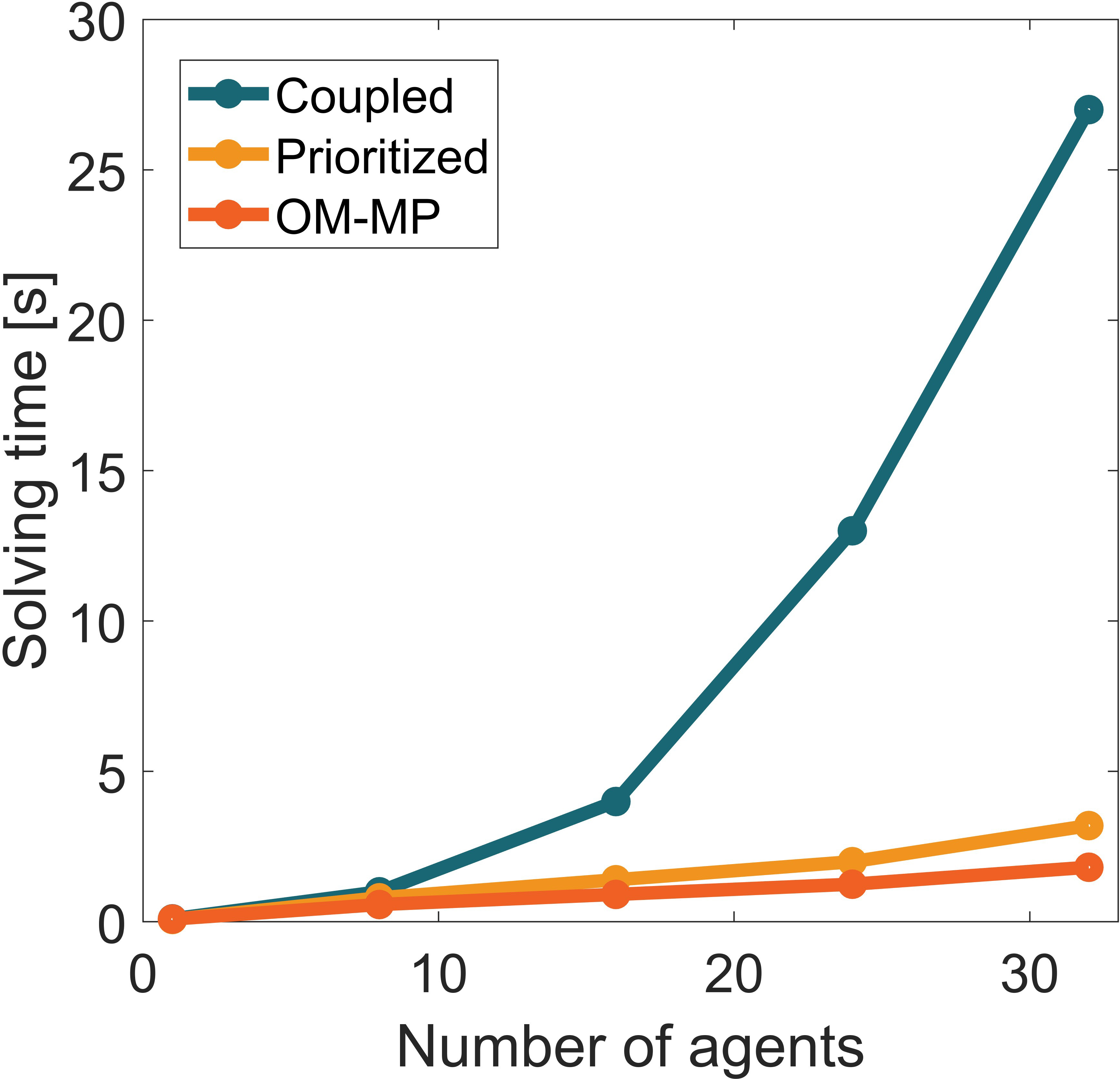}
        \label{com_fig_a}
    }
    \quad
    \subfigure[]{
        \includegraphics[width=0.445\columnwidth]{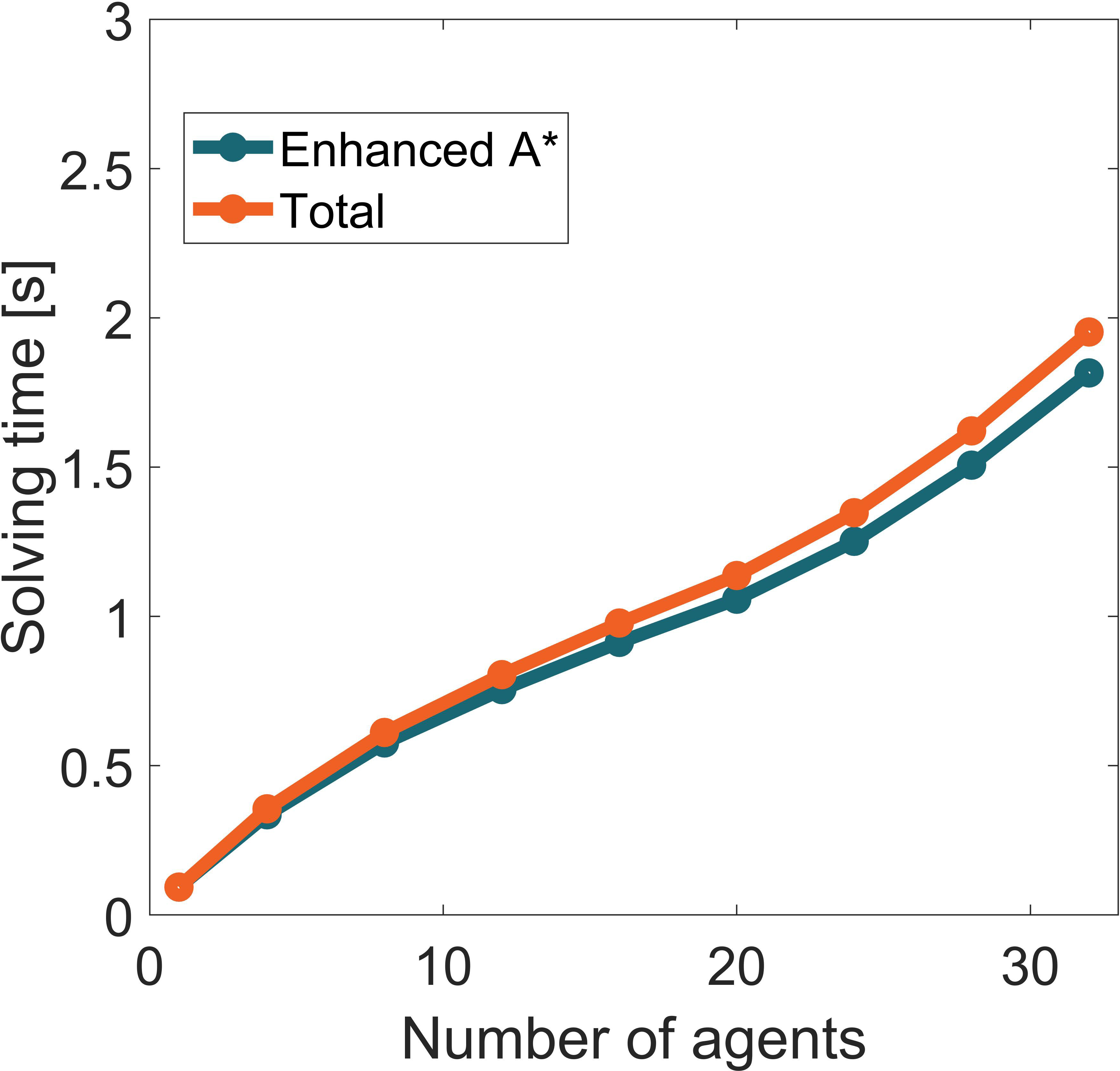}
        \label{com_fig_b}
    }
    \caption{Comparison of the OM-MP approach and coupled, and prioritized method in reference \cite{li2020efficient}. }
    \label{comparison}
\end{figure}

In the context of the warehouse scenario, we evaluate the performance of the OM-MP approach. The warehouse map is designed with dimensions of 500 $\times$ 300 pixels, corresponding to a real-world size of 25 m ${\times}$ 15 m. Start and goal positions are evenly distributed on the left and right sides of the map. The average time consumption for different numbers of robots is assessed based on 10 experiments, and the results are presented in Table \ref{table3} and Fig~\ref{com_fig_b}. The global planner solving time for OM-MP demonstrates an almost linear increase relative to the number of robots. After this initial linear period, each robot can execute its individual actions.
\par
The original complex trajectory optimization problem can be referred to as a coupled trajectory optimization problem involving multiple robots \cite{li2020efficient}. As depicted in Fig~\ref{com_fig_a}, when dealing with 32 robots, the direct solution requires around 27 seconds, the simple priority strategy takes approximately 5 seconds, and the OM-MP approach commences begins movement towards the target positions in less than 2 seconds. This comparison underscores the efficiency of the OM-MP strategy in rapidly generating feasible paths for numerous robots.

\subsection{Queuing scenario}
\begin{figure}[h]
    \centerline{\includegraphics[width=\columnwidth]{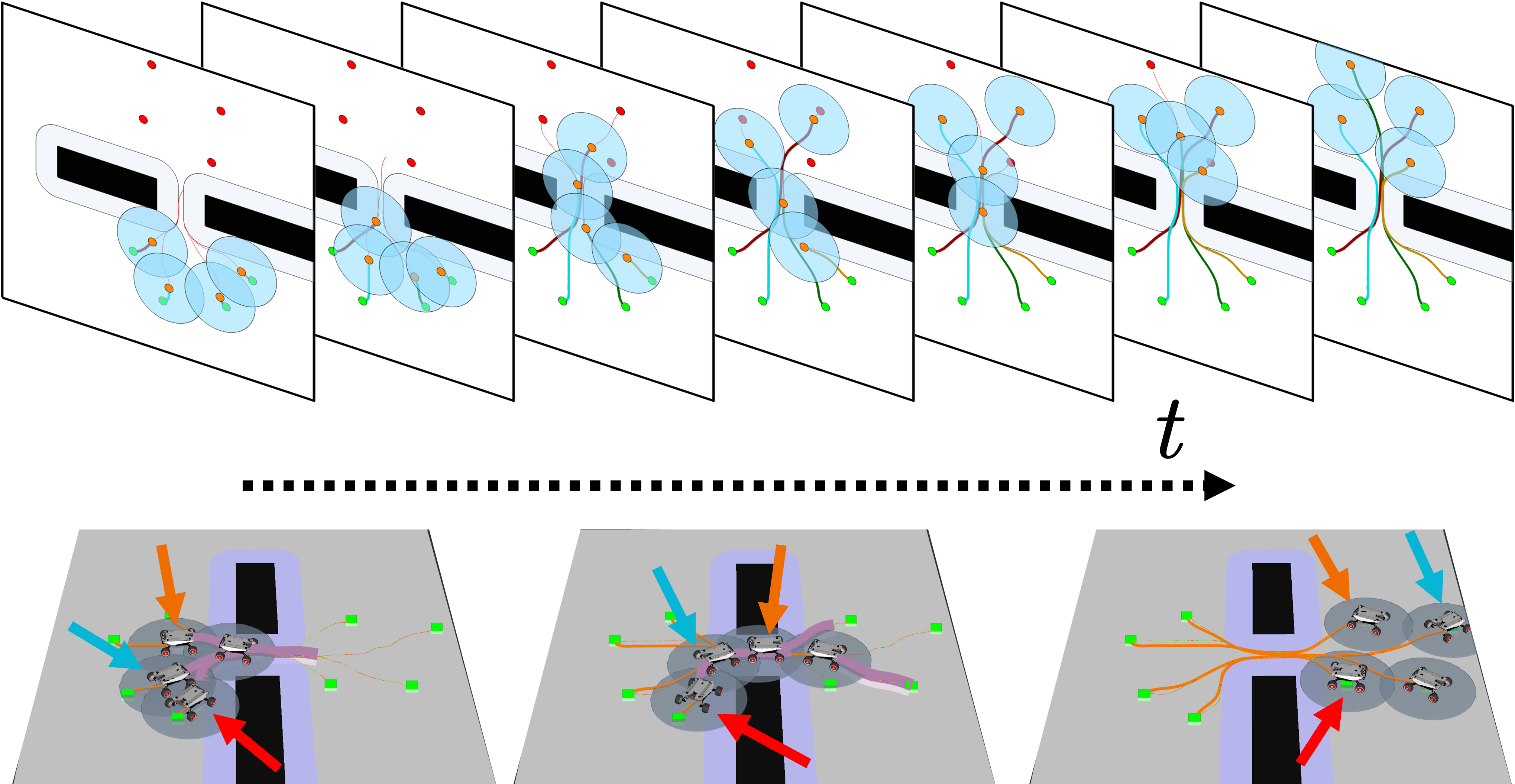}}
    \caption{The demonstration of queuing scenario, the up image shows the simulation results in ROS environment, which has been imported into Matlab for visualization. Black area represent obstacles, while light blue circles indicate the communication range of the intelligent robot. The bottom figure shows a key frame screenshot of the simulation in Rviz, from left to right representing the entire trajectory planning process.}
    \label{LINE_com}
\end{figure}

The scenario where robots need to queue up to navigate through a narrow passage underscores the significance of employing a hierarchical priority approach for achieving overall optimally, as illustrated in Fig~\ref{LINE_com}. In this scenario, the red arrow points to the robot closer to the intersection compared to the blue arrow. However, due to its proximity to its destination, its weights is lower. Consequently, during local planning, the trajectory planned for the red arrow should behind that of the blue arrow's, potentially resulting in a slower pace or even a halt for the red arrow robot until the blue arrow's robot passes through the conflict area. Similarly, the orange arrow designates the robot with a lower weight (due to its shorter residual path compared to the blue one). Nonetheless, the orange one is closer to the intersection, it can pass through earlier while adhering to speed constraints. Hence, the hierarchical weight-first strategy prioritizes extending the length of the shorter queue in what can be referred to as the "barrel effect" (where longer paths are planned first). This strategy aims to minimize the running time of the robot with longer distance to achieve more favorable global characteristics.

\subsection{Replan scenario}
\begin{figure}[h]
    \centerline{\includegraphics[width=\columnwidth]{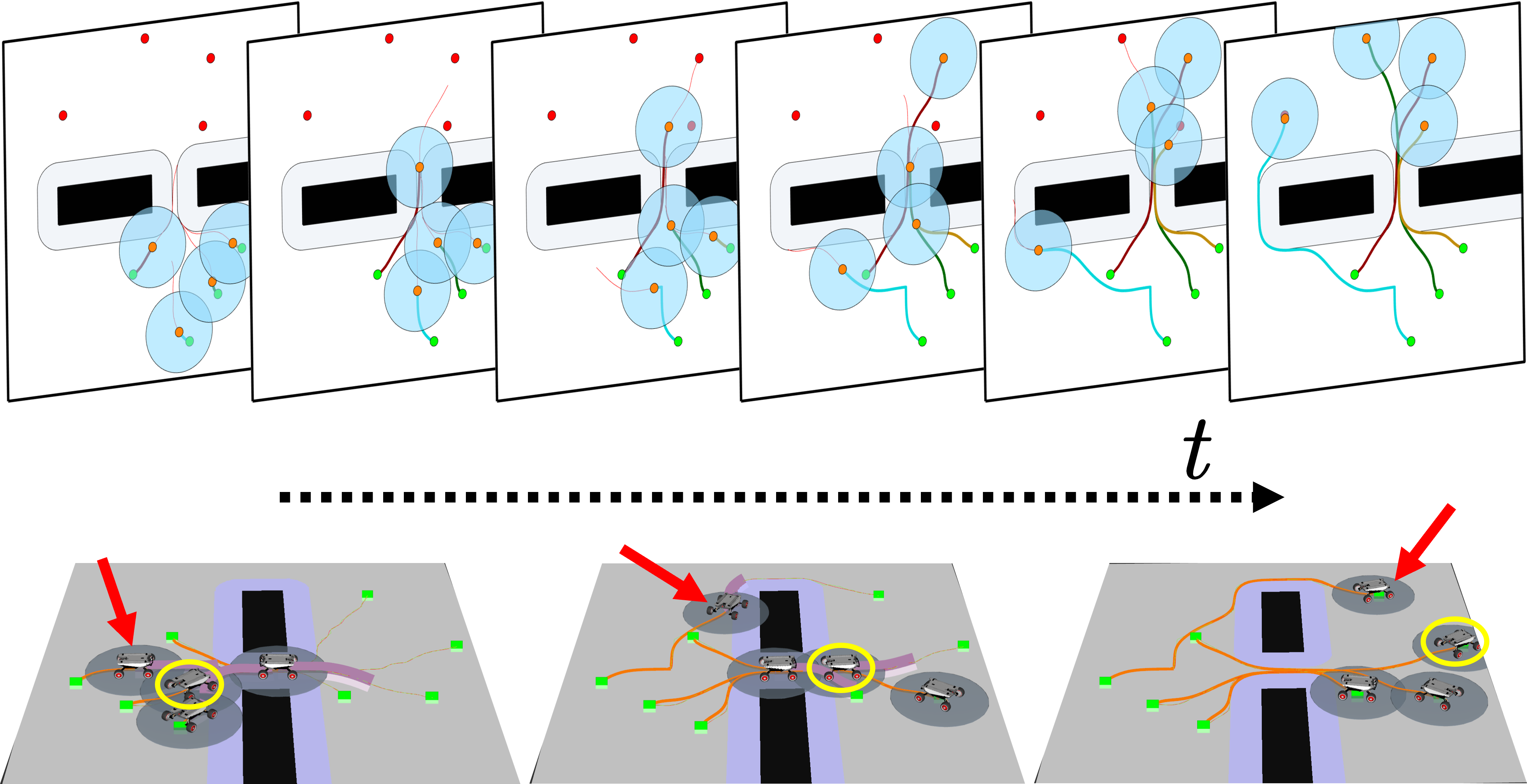}}
    \caption{The illustration of the simulation scenario when a certain intelligent robot is performing replanning.}
    \label{replan_com}
\end{figure}

When multiple robots converge at a narrow intersection simultaneously, lower-priority robots may experience prolonged waiting times. In our approach, the replanning strategy is introduced to tackle this issue, as depicted in Fig~\ref{replan_com} and Algorithm \ref{algorithm3}. In this scenario, the red arrow indicates the robot with lower priority compared to the robots within the yellow circle. When the locally planned trajectory of the lower-priority robot with a longer time $t$ extends a specified threshold, it initiates a replanning request to the global planner. Upon receiving the new path, the robot conducts a comparison between the travel time of the new path, considering a reference speed and the current path, with that of the current path, as shown in Algorithm \ref{algorithm3}. Then, the local planner selects the shorter of the two paths to track. The middle image at the bottom of Fig~\ref{replan_com} provides an illustrative example of this situation.
\subsection{Scenarios analyst}
\par
\begin{figure}[h]
    \centering
    \subfigure[Warehouse scenario]{
        \includegraphics[width=0.44\columnwidth]{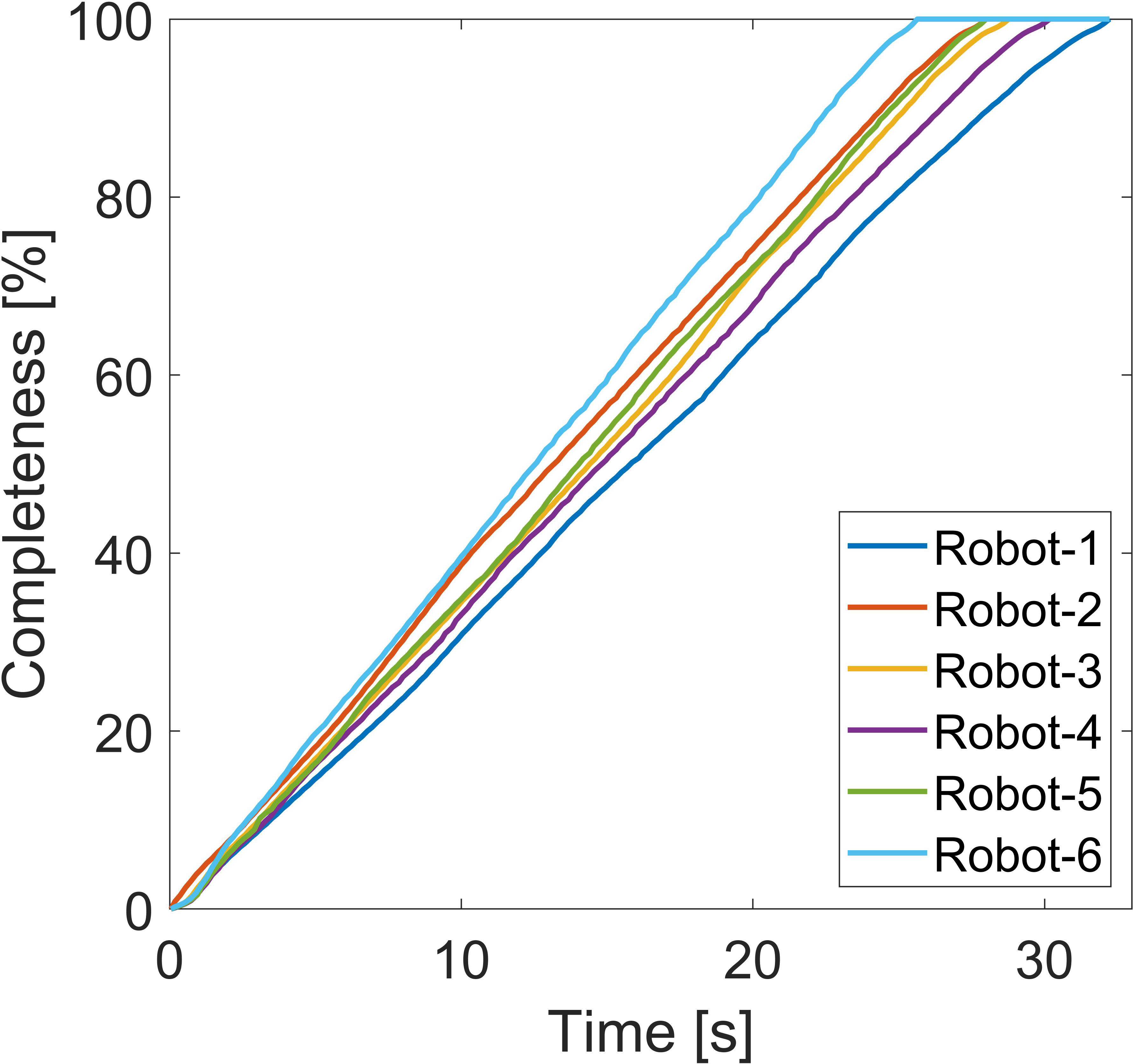}
        \label{com_fig:a}
    }
    \quad
    \subfigure[Random scenario]{
        \includegraphics[width=0.44\columnwidth]{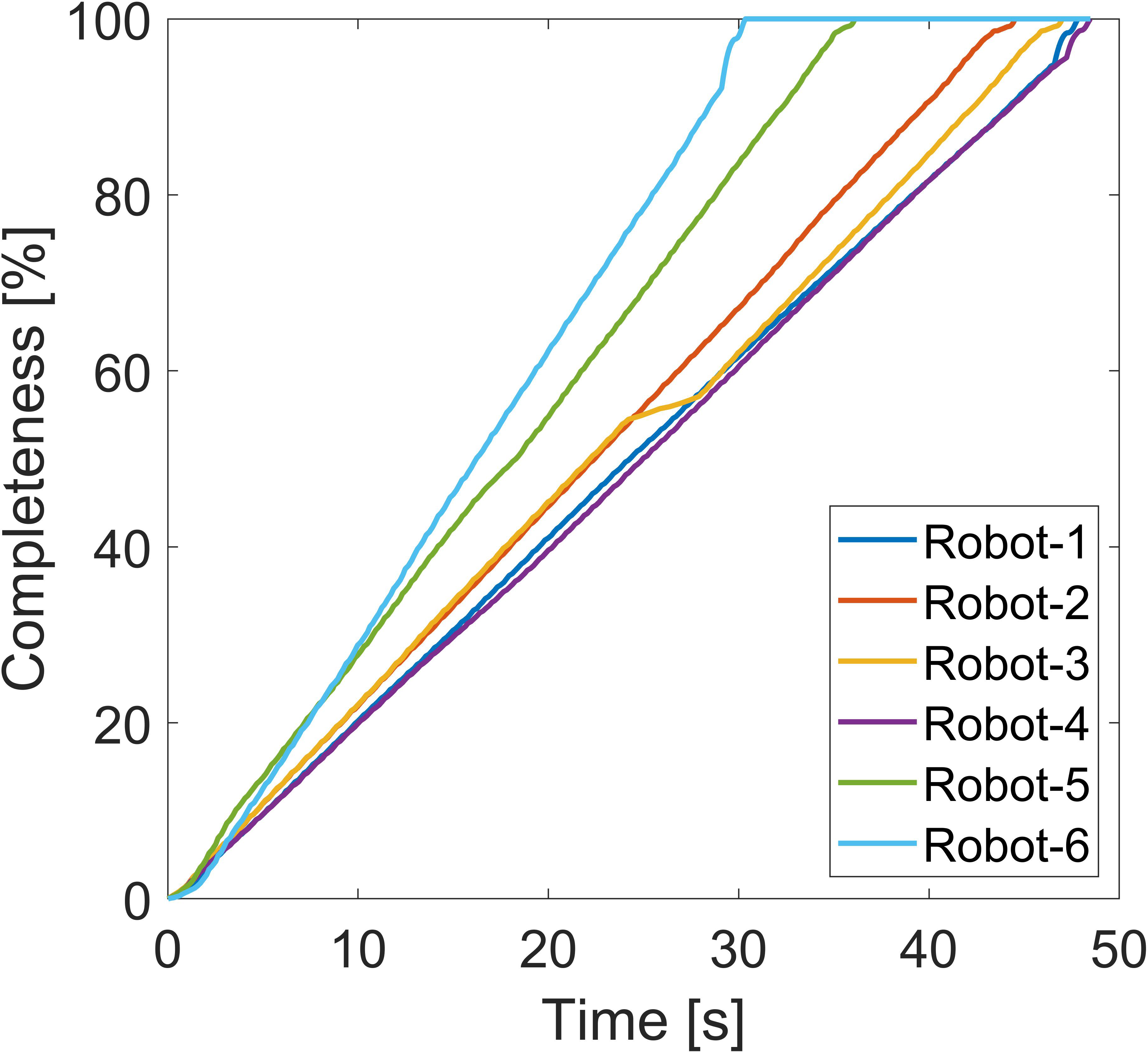}
        \label{com_fig:b}
    }
    \quad
    \subfigure[Queuing scenario]{
        \includegraphics[width=0.44\columnwidth]{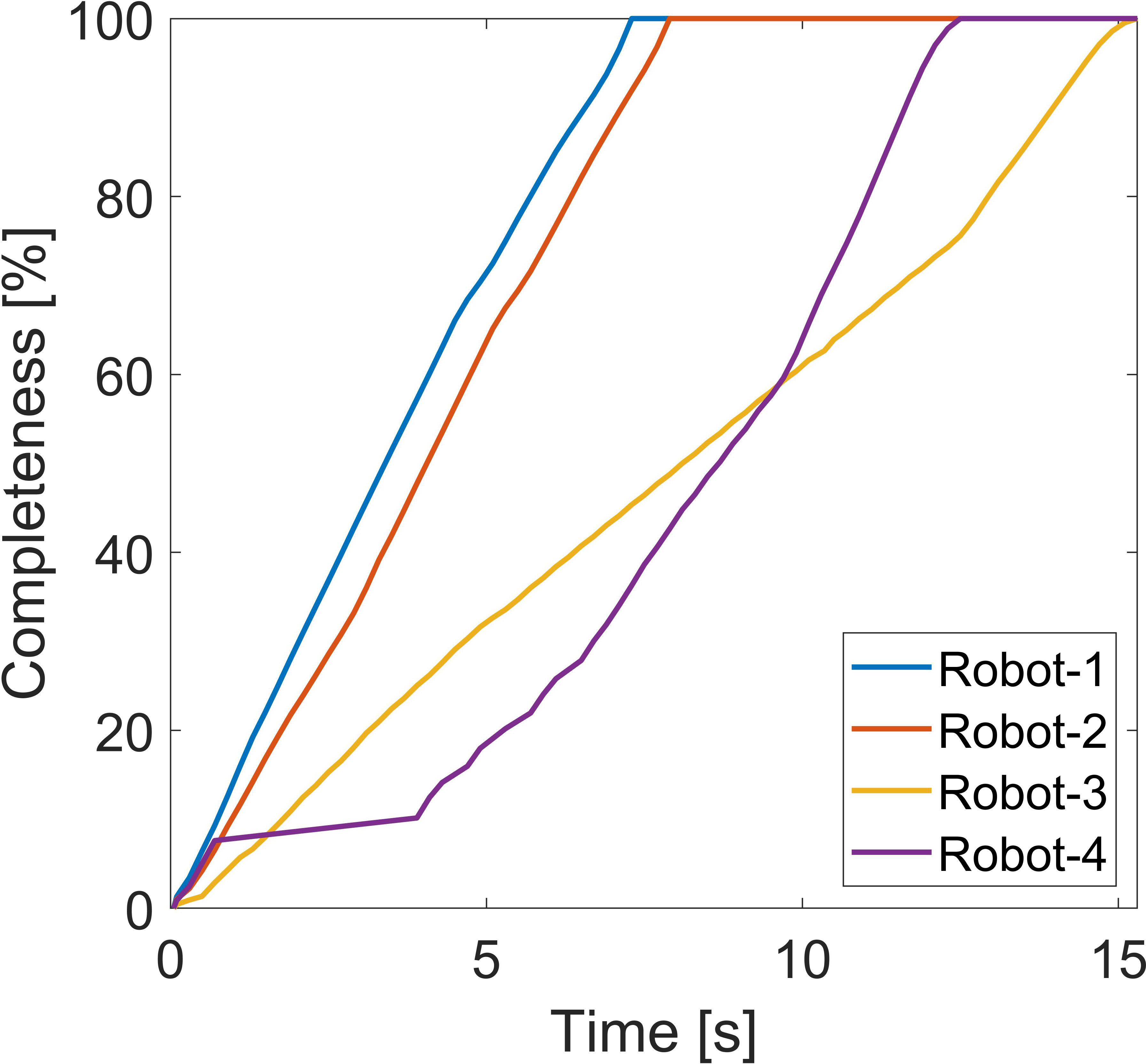}
        \label{com_fig:c}
    }
    \quad
    \subfigure[Replan scenario]{
        \includegraphics[width=0.44\columnwidth]{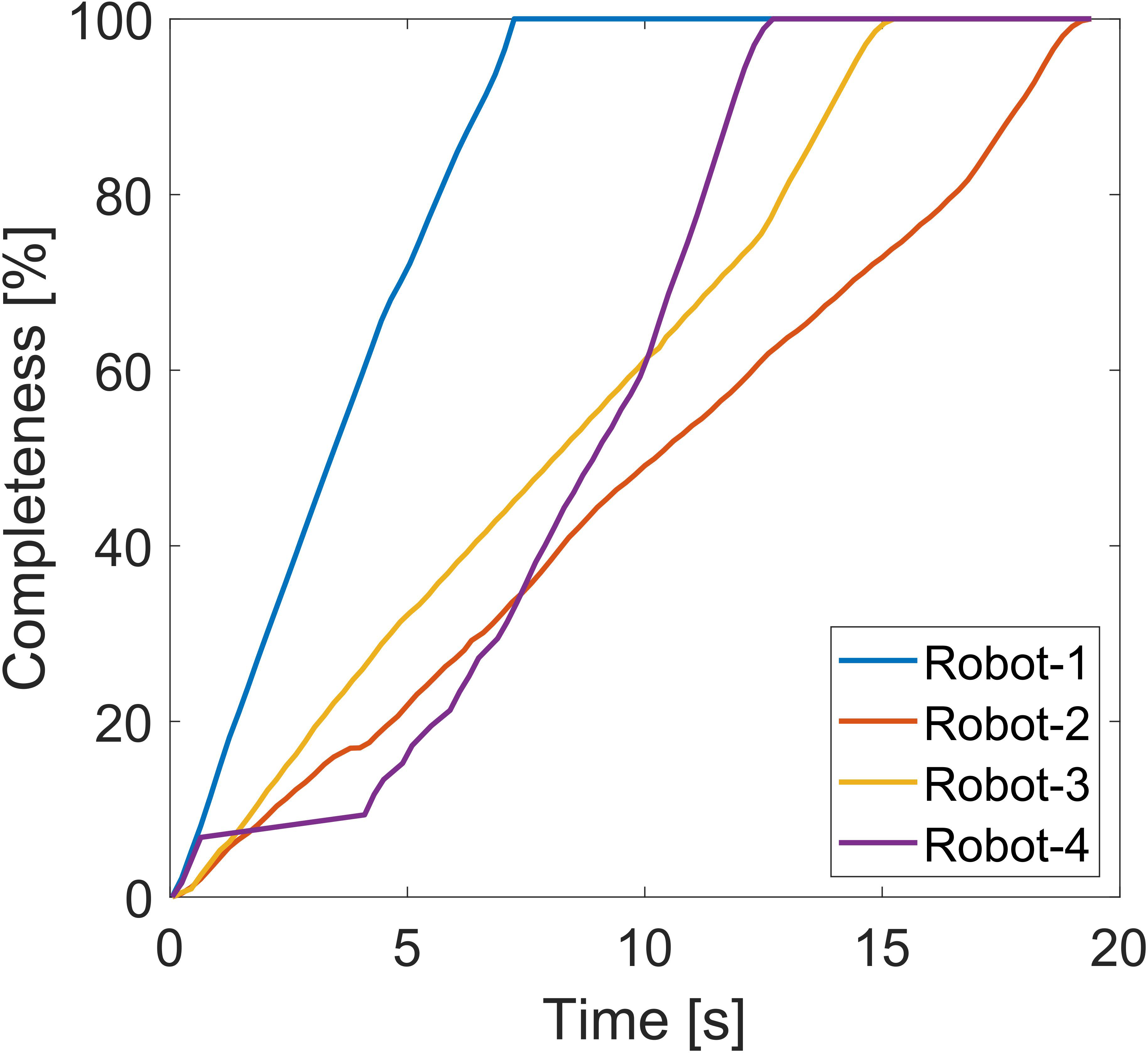}
        \label{com_fig:d}
    }
    \caption{Demonstration of the overall completion in four simulated scenarios.}
    \label{completeness}
\end{figure}
\par
Fig~\ref{completeness} illustrates the overall completion of four scenarios. It is evident that the robot with the longest path (robot-1 in \subref{com_fig:a}, robot-4 in \subref{com_fig:b}, robot-4 in \subref{com_fig:c}, and robot-4 in \subref{com_fig:d} of Fig~\ref{completeness}) remains virtually undisturbed by other external influences and continues to move steadily at its own speed, ensuring optimal overall task fulfillment in each scenario. 
\par
In \subref{com_fig:c} of Fig~\ref{completeness}, robot-4 experiences a noticeable slowdown and even comes to a halt during the interval of approximately 1.5-4 seconds. This is attributed to the higher-priority robot-3 obstructing its movement, as indicated by the red and blue arrows in Fig~\ref{LINE_com}. Similar situations also occur with robot-3 in \subref{com_fig:b} and robot-4 in \subref{com_fig:d}.
\par
\begin{figure}[h]
    \centering
    \subfigure[]{
        \includegraphics[width=0.46\columnwidth]{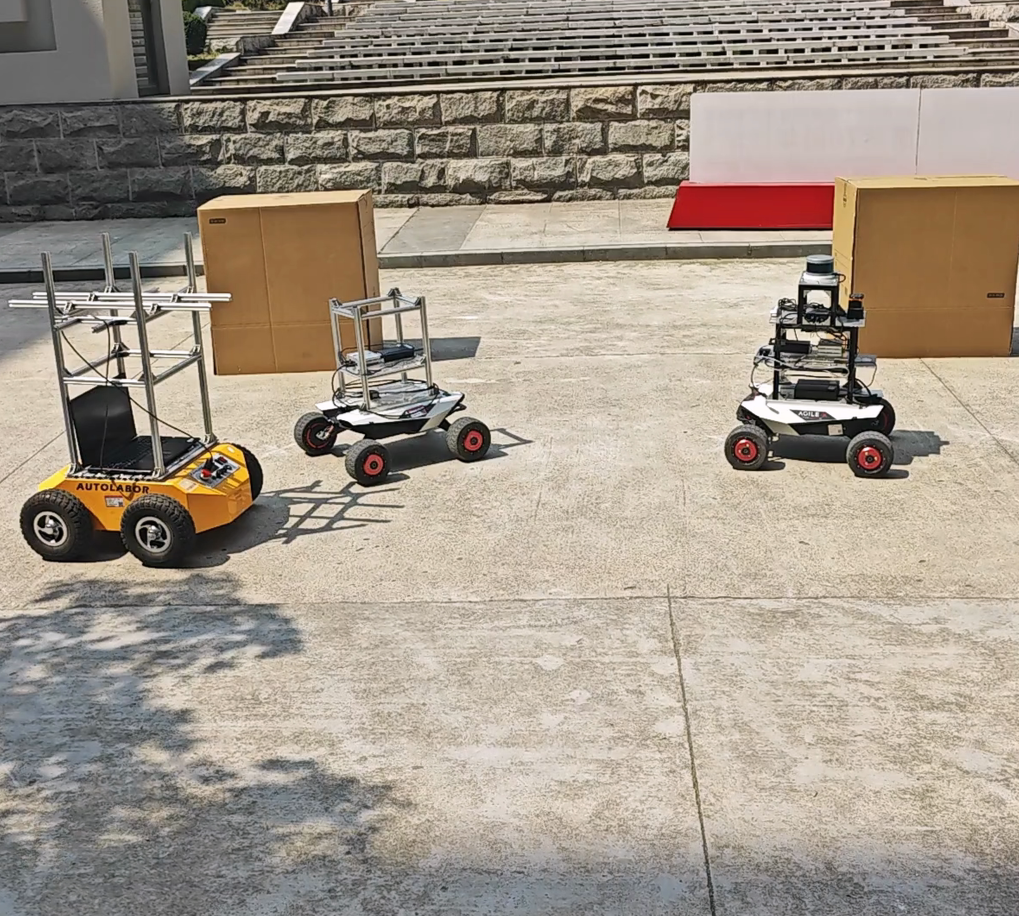}
        \label{real_exp}
    }
    \quad
    \subfigure[]{
        \includegraphics[width=0.42\columnwidth]{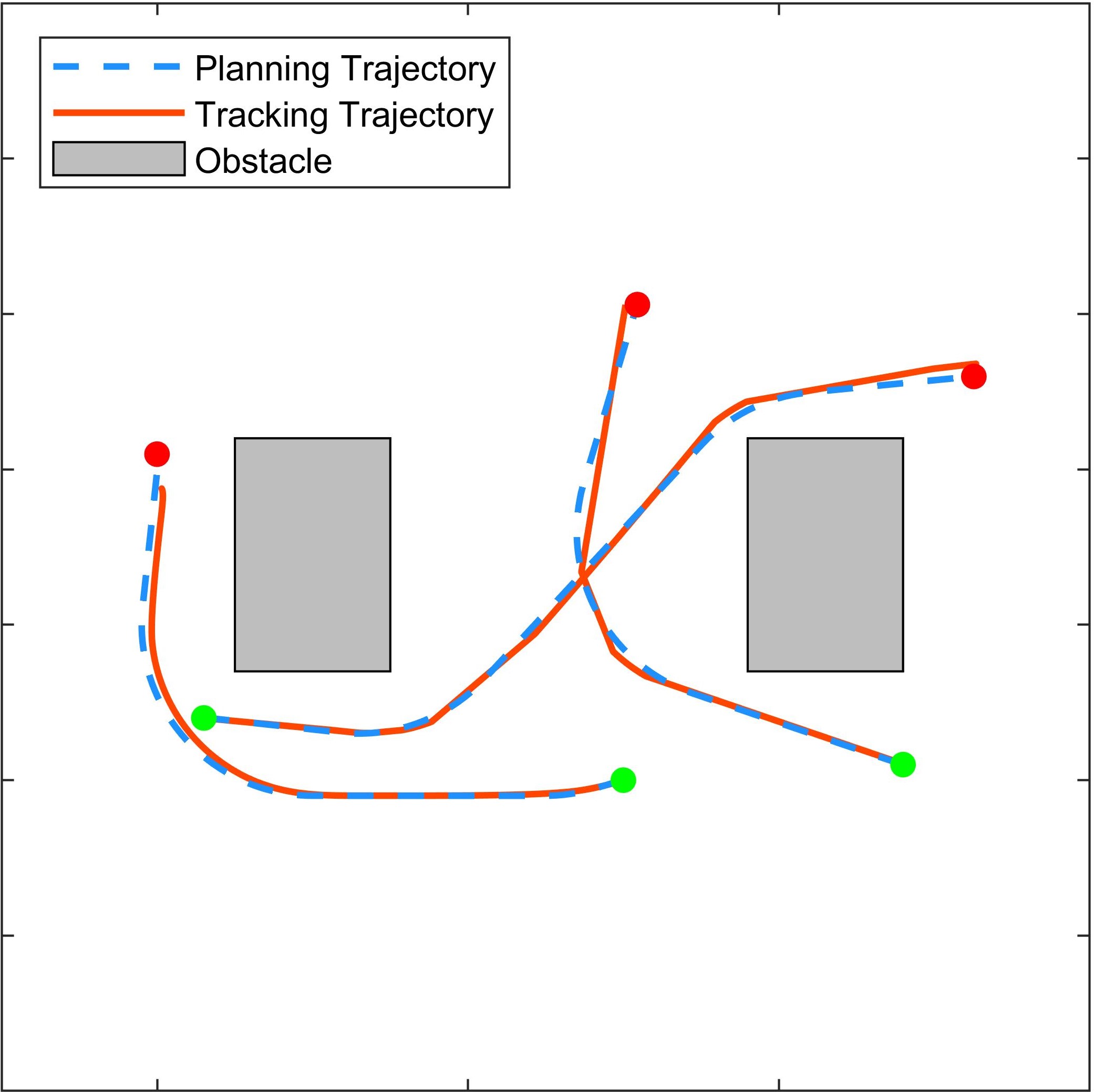}
        \label{sim_exp}
    }
    \caption{The demonstration of the OM-MP framework experiment of replanning and queuing scenarios in HUST.}
    \label{experiment}
\end{figure}

The OM-MP framework is illustrated in on three mobile robots (as shown in Fig~\ref{experiment}), which is equipped with the Jetson NX (in AGILE$\cdot$X, the white two) and a laptop (in Autolabor, the yellow one) as the local planner, and another as the global planner.

\section{Discussion \&\& Conclusion}
In this work, we proposed a real-time framework for multi-robot path planning and trajectory optimization and control. We introduced AGS to build safe zones, and NURBS curves were used in global path planning to ensure path smoothness. A hierarchical optimization architecture was adopted, with a replanning algorithm enhanced for individual robots to ensure the suboptimally guarantee of single robot paths. Additionally, the local planner optimization utilized a  communication-based approach in radius $R$ to reduce the complexity of optimization and ensure solution efficiency. Moreover, the framework prioritized the planning of robots with longer paths using a weight-based priority strategy in conflict zones  and employed replanning algorithms to converge to a globally optimal solution.
\par
Compared to Multi-Agent Path Finding (MAPF) algorithms such as CBS and ECBS, our OM-MP algorithm offers several significant advantages. Notably, it eliminates the need for a separate planning time slot before robot movement begins. Instead, our approach swiftly obtains solutions using the AGS algorithm and fine-tunes them in real-time as robots navigate, ensuring efficient and rapid planning. Additionally, the two-layer architecture we've employed offers distinct benefits. The upper layer is primarily responsible for search and coordinate operations, covering both initial planning and subsequent re-planning for all robots. Meanwhile, the lower layer focuses on specific local trajectory optimization and actuator control signal. This design can be seen as a semi-clustered architecture, where the planning complexity of each robot primarily increases within the upper layer. Each robot independently computes its local trajectory, significantly enhancing the efficiency of solving extensive multi robot problems when contrasted with CBS and similar algorithms that emphasize global planning strategies.
\par
Furthermore, the OM-MP algorithm has the ability to dynamically adjust control inputs based on each robot's movement state and tracking conditions, a feature not present in CBS and comparable algorithms. This adaptability renders the OM-MP algorithm suitable not only for multi robots planning but also for planning intelligent connected vehicles in specific areas such as parks and intersections.
\par
However, it's important to acknowledge that while the OM-MP algorithm architecture has demonstrated efficiency across scenarios described above, there are still certain limitations to consider. For instance, in scenarios featuring multiple intersections requiring queuing, the constraints of the re-planning algorithm might cause robot repeatedly oscillate between multiple intersections, potentially deviating from the global optimal solution. However, such extreme and rare situations are not thoroughly analyzed in this context.

\par
\appendices
\section{Definition of Equation~\ref{state_trans} and Equation~\ref{affine_dynamics}}
\label{app:A}
\subsection{System~\ref{state_trans}}

The system described by Equation~\ref{state_trans} is given by:

\begin{equation*}
    \dot{q} = Aq + BU
\end{equation*}

where the matrices $A$ and $B$ are defined as:

\[
A = \begin{bmatrix}
    0 & 0 & \frac{\cos\theta}{2} & \frac{\cos\theta}{2} & 0 & 0 \\
    0 & 0 & \frac{\sin\theta}{2} & \frac{\sin\theta}{2} & 0 & 0 \\
    0 & 0 & 0 & 0 & 0 & 0 \\
    0 & 0 & 0 & 0 & 0 & 0 \\
    0 & 0 & \frac{1}{l_w} & -\frac{1}{l_w} & 0 & 0 \\
    0 & 0 & 0 & 0 & 0 & 1 \\
\end{bmatrix}
\]

and

\[
B = \begin{bmatrix}
    0 & 0 \\
    0 & 0 \\
    r & 0 \\
    0 & r \\
    0 & 0 \\
    0 & 0 \\
\end{bmatrix}
\]

\subsection{System~\ref{affine_dynamics}}

The system described by Equation~\ref{affine_dynamics} is given by:

\begin{equation*}
    \dot{s} = f(s) + g(s) u
\end{equation*}

where the functions $f(s)$ and $g(s)$ are defined as:

\[
f(s) = \begin{bmatrix}
    \frac{(v_r + v_l)\cos\theta}{2} \\
    \frac{(v_r + v_l)\sin\theta}{2} \\
    0 \\
    0 \\
    \frac{v_r - v_l}{l_w} \\
    1 \\
\end{bmatrix}
\]

and

\[
g(s) = \begin{bmatrix}
    0 & 0 \\
    0 & 0 \\
    r & 0 \\
    0 & r \\
    0 & 0 \\
    0 & 0 \\
\end{bmatrix}
\]

\section{Designing the Extended Class $\mathcal{K}_\infty$ Function, $\alpha$}
\label{app:B}
In this section, we present the design of the extended class $\mathcal{K}_\infty$ function, denoted as $\alpha$, for a safe set $\mathcal{C}$ that satisfies the definition presented in Equation~\ref{safe_set_def}. Additionally, we consider a Control Barrier Function (CBF) denoted as $\mathcal{B}(s)$ for the system $\mathcal{S}$, such that any feasible planning point $s$ within $\mathcal{S}$ adheres to the following constraints:

\begin{equation*}
\begin{cases}
    \mathcal{B}(s) > 0, \quad s \in \mathrm{Int}(\mathcal{C})\\
    \mathcal{B}(s) = 0, \quad s \in \partial(\mathcal{C})
\end{cases}
\end{equation*}

The satisfaction of CBFs, where $\mathcal{B}(s) \geqslant 0$, ensures the safety of the planning. We aim to find a boundary function $\mathcal{A}(s) > 0$ such that $\mathcal{B}(s) \geqslant \mathcal{A}(s) > 0$.

Considering:

\begin{equation*}
\dot{\mathcal{A}} = -\gamma \mathcal{A}, \quad \gamma > 0
\end{equation*}

This leads to:

\begin{equation*}
\mathcal{A}(t) = e^{-\gamma t}\mathcal{A}_0 > 0
\end{equation*}

Consequently, we have $\mathcal{B}(s) = \mathcal{A}(s) = e^{-\gamma s}\mathcal{A}_0 > 0$, ensuring collision-free planning.

Next, we consider:

\begin{equation*}
\begin{aligned}
    \dot{\mathcal{B}}(s) &= \mathcal{L}_f\mathcal{B}(s) + \mathcal{L}_g\mathcal{B}(s)u \\
    &= \dot{\mathcal{A}}(s) = -\gamma \mathcal{A}(s) = -\gamma \mathcal{B}(s) > 0
\end{aligned}
\end{equation*}

We can rewrite the above equation as:

\begin{equation*}
\dot{\mathcal{B}} + \gamma \mathcal{B}(s) > 0
\end{equation*}

Therefore, the extended class $\mathcal{K}_\infty$ function $\alpha(s)$ is defined as:

\begin{equation*}
\alpha(s) = \gamma \mathcal{B}(s)
\end{equation*}

\section{Notes and Our future work}
In our simulation experiments, our initial goal was to conduct tracking simulations of physical objects using Gazebo. However, we encountered a challenge related to Gazebo's performance when loading multiple model files. Specifically, Gazebo's execution speed was considerably slower (less than 1/10th) compared to the real environment, especially when dealing with multiple model files. This disparity posed difficulties in achieving alignment between Gazebo-based simulations and the global computing results. Consequently, we opted to rely solely on Rviz for our simulation demonstrations. Given that our architecture follows a distributed computing approach, each agent requires real-time computing resources. This configuration has implications in simulation, because each agent needs to create its own independent thread, and each thread needs to perform high frequency of NLP computation, however, this situation does not happen in practice, as each agent is individually equipped with a processing unit. 

\bibliographystyle{IEEEtran}

\bibliography{ref}

\end{document}